%% file: main.tex
\PassOptionsToPackage{normalem}{ulem}

\documentclass[]{TEAI}

\usepackage{helvet}
\usepackage[utf8]{inputenc}
\usepackage[T1]{fontenc}

\usepackage{amsmath,amssymb,mathtools,amsfonts}
\usepackage{graphicx}
\usepackage{subcaption}
\usepackage{wrapfig}
\usepackage{caption}

\usepackage{booktabs}
\usepackage{multirow}
\usepackage{siunitx}
\usepackage{enumitem}
\usepackage{xspace}
\usepackage{listings}

\usepackage{xcolor}         
\usepackage{hyperref}
\usepackage{url}

\usepackage{natbib}

\usepackage{algorithm}
\usepackage{algpseudocode}
\usepackage{catchfile}

\makeatletter
\@ifpackageloaded{ulem}{}{
  \usepackage[normalem]{ulem}
}
\makeatother

\usepackage{tcolorbox}
\tcbuselibrary{breakable,skins}

\providecommand{\email}[1]{\href{mailto:#1}{#1}}

\colorlet{highlight}{cyan!10}
\definecolor{gaincolor}{RGB}{0,128,0}
\definecolor{losscolor}{RGB}{180,60,60}
\definecolor{bestcolor}{RGB}{255,245,220}

\lstdefinestyle{case}{
  basicstyle=\ttfamily\scriptsize,
  columns=fullflexible,
  keepspaces=true,
  showstringspaces=false,
  breaklines=true,
  breakatwhitespace=false,
  upquote=true,
  aboveskip=2pt,
  belowskip=2pt,
  xleftmargin=0.6em
}
\newtcolorbox{casebox}[2][]{%
  enhanced,
  breakable,
  width=\columnwidth,
  colback=gray!4,
  colframe=black!25,
  colbacktitle=black!80,
  coltitle=white,
  fonttitle=\bfseries\small,
  title=#2,
  boxrule=0.45pt,
  arc=2mm,
  left=1.2mm,right=1.2mm,top=0.9mm,bottom=0.9mm,
  fontupper=\footnotesize,
  fontlower=\footnotesize,
  pad at break*=1mm,
  before skip=8pt,
  after skip=6pt,
  segmentation style={solid,black!20},
  #1
}

\input{macros}

\title{NEX: Neuron Explore–Exploit Scoring for Label-Free Chain-of-Thought Selection and Model Ranking}

\author{
Kang Chen\textsuperscript{1*},
Zhuoka Feng\textsuperscript{1*},
Sihan Zhao\textsuperscript{1},
Kai Xiong\textsuperscript{},
Junjie Nian\textsuperscript{1}\\
Yaoning Wang\textsuperscript{1},
Changyi Xiao\textsuperscript{1},
Yixin Cao\textsuperscript{1,2$\dagger$}
}

\affiliation[1]{\mbox{Fudan University}}
\affiliation[2]{\mbox{Shanghai Innovation Institute}}

\correspondence{\email{Kchen24@m.fudan.edu.cn}\quad \email{yxcao@fudan.edu.cn}}
\abstract{
\input{sections/abstract}
}
\begin{document}
\maketitle

% Footnotes under title
\begingroup
\renewcommand{\thefootnote}{}
\footnotetext{* Contributed equally.}
% \footnotetext{** Contributed equally (Co-second authorship).}
\footnotetext{$\dagger$ Corresponding author.}
\endgroup

\input{sections/introduction}
\input{sections/related_work}
\input{sections/method}
\input{sections/experimental_setup}
\input{sections/results_fzk}

\input{sections/limitations}
\input{sections/conclusion}

\clearpage
\bibliographystyle{plainnat}
\bibliography{references}

\appendix
\input{sections/appendix}

\end{document}

%% file: macros.tex
\newcommand{\NEX}{\textsc{NEX}\xspace}
\newcommand{\eg}{e.g.\xspace}

\newcommand{\figref}[1]{Figure~\ref{#1}}
\newcommand{\tabref}[1]{Table~\ref{#1}}
\newcommand{\secref}[1]{Section~\ref{#1}}

%% file: sections/introduction.tex
\section{Introduction}

Large language models~(LLMs) have made rapid progress on tasks requiring multi-step reasoning
\citep{wei2022chain,lewkowycz2022minerva},
yet state-of-the-art performance on hard problems often depends on how much compute is allocated at inference time
\citep{snell2025ttcoptimal,zhang2025ttssurvey}, a paradigm commonly referred to as test-time scaling.
Different from single-pass decoding, test-time scaling elicits chain-of-thought~(CoT) traces and spend additional compute
by sampling, searching, or verifying multiple candidates before selecting a final output
\citep{wang2023selfconsistency,yao2023tree}.
This paradigm shifts the bottleneck from generation to selection
\citep{welleck2024metageneration,zhang2025ttssurvey},
often in settings where labeled validation data on the target distribution is scarce or unavailable and selection must
rely on self-supervised signals \citep{kadavath2022lmsknow,kuhn2023semantic}.
At the same time, such methods often require proper exploration capability, thus simply increasing thinking is not always beneficial: more stochasticity or longer CoTs can induce
wandering, repetition, or diminishing returns under fixed compute budgets
\citep{holtzman2020curious,wu2024inferencescaling}.
These observations raise a basic question: what constitutes productive exploration during reasoning, and how can we
detect it without access to answers?

% Modern reasoning LMs are increasingly produced not as a single checkpoint but as a \emph{family of closely related variants}: merged checkpoints along a merge coefficient, instruction--thinking mixtures trained with different data ratios, or other style-mixture mechanisms.
% In such settings, the practical question is no longer only ``how to train a good model,'' but also \textbf{how to choose} among many near-siblings~(\eg Which reasoning trace should we trust at inference time?)
% (1)~\emph{Which model variant should we deploy?}
% (2)~\emph{Which reasoning trace (CoT) should we trust at inference time?}
% (3)~\emph{Which generated training samples should we keep to train a smaller student?}
% Achieving this with labeled evaluation is expensive and often mismatched to the target distribution
% \citep{welleck2024metageneration,zhang2025ttssurvey}.

To delve into this, we study reasoning through the lens of the exploration--exploitation trade-off \citep{sutton2018rl,yao2023tree}.
To vary the model's exploration tendency, we start from two endpoint models with distinct behaviors:
an Instruct model that tends to explore too little and a Thinking model that tends to explore too much.
Using parameter-space model merging and interpolation \citep{wortsman2022soups,yadav2023ties}, we
generate a family of intermediate model variants that span a continuum of behaviors and sample CoT traces from each
variant.
Across multiple benchmarks, we find that accuracy as a function of an entropy-based exploration proxy follows a clear
inverted-U pattern (\figref{fig:pair}a), implying that there exists an optimal balance between exploration and exploitation and that
excessive exploration can be redundant or harmful.

\usetikzlibrary{calc,positioning}

\captionsetup[subfigure]{labelformat=parens,labelsep=space}

\begin{figure*}[t]
\centering
\begin{tikzpicture}

% (a) 左侧图
\node[inner sep=0] (A) {%
  \begin{subfigure}[b]{0.32\textwidth}
    \centering
    \includegraphics[width=\linewidth]{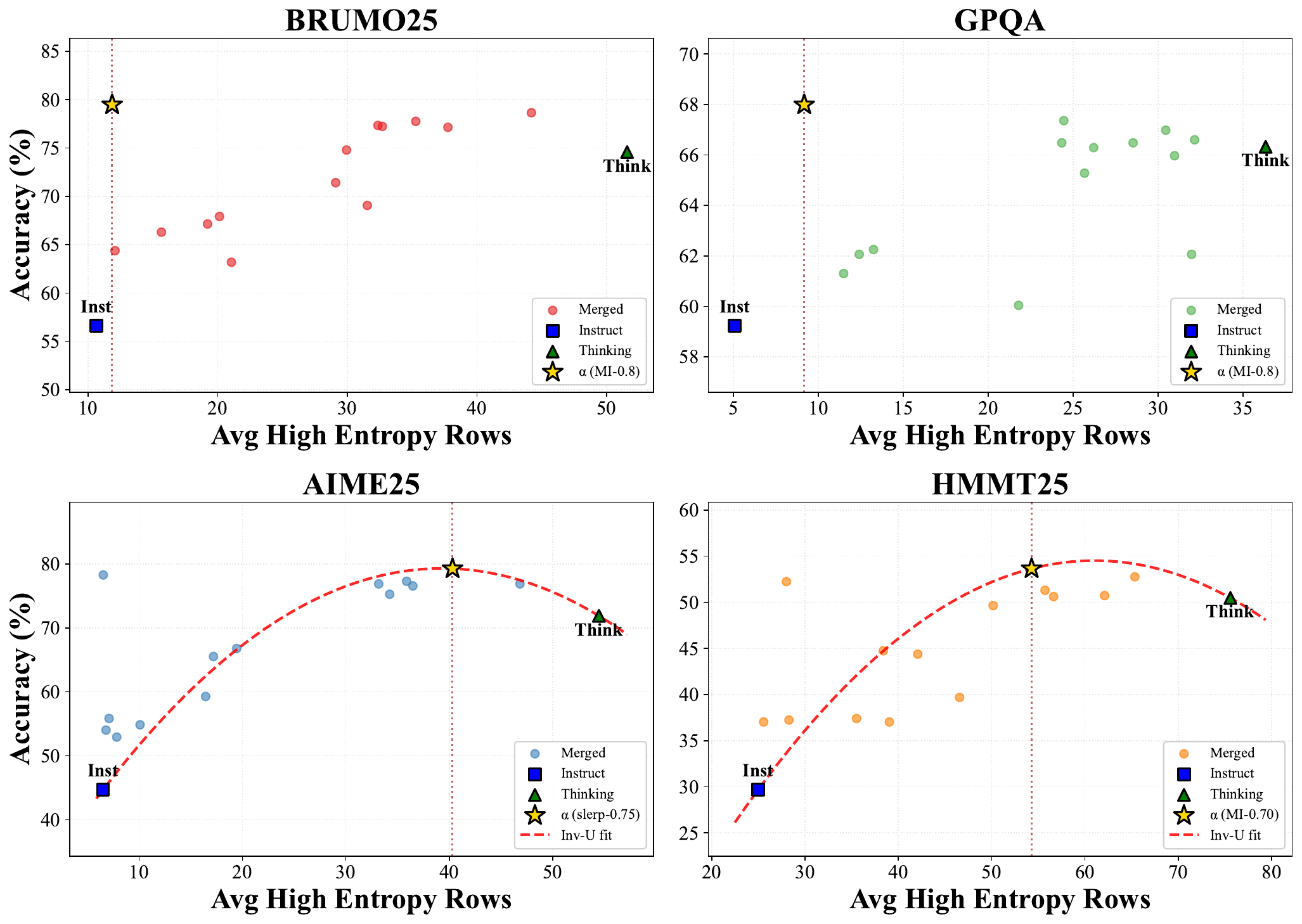}
    \caption{}\label{fig:explore_inverted_u}
  \end{subfigure}
};

% (b) 中间新加入的图
\node[inner sep=0, right=0.01\textwidth of A] (B) {%
  \begin{subfigure}[b]{0.32\textwidth}
    \centering
      \includegraphics[width=\linewidth]{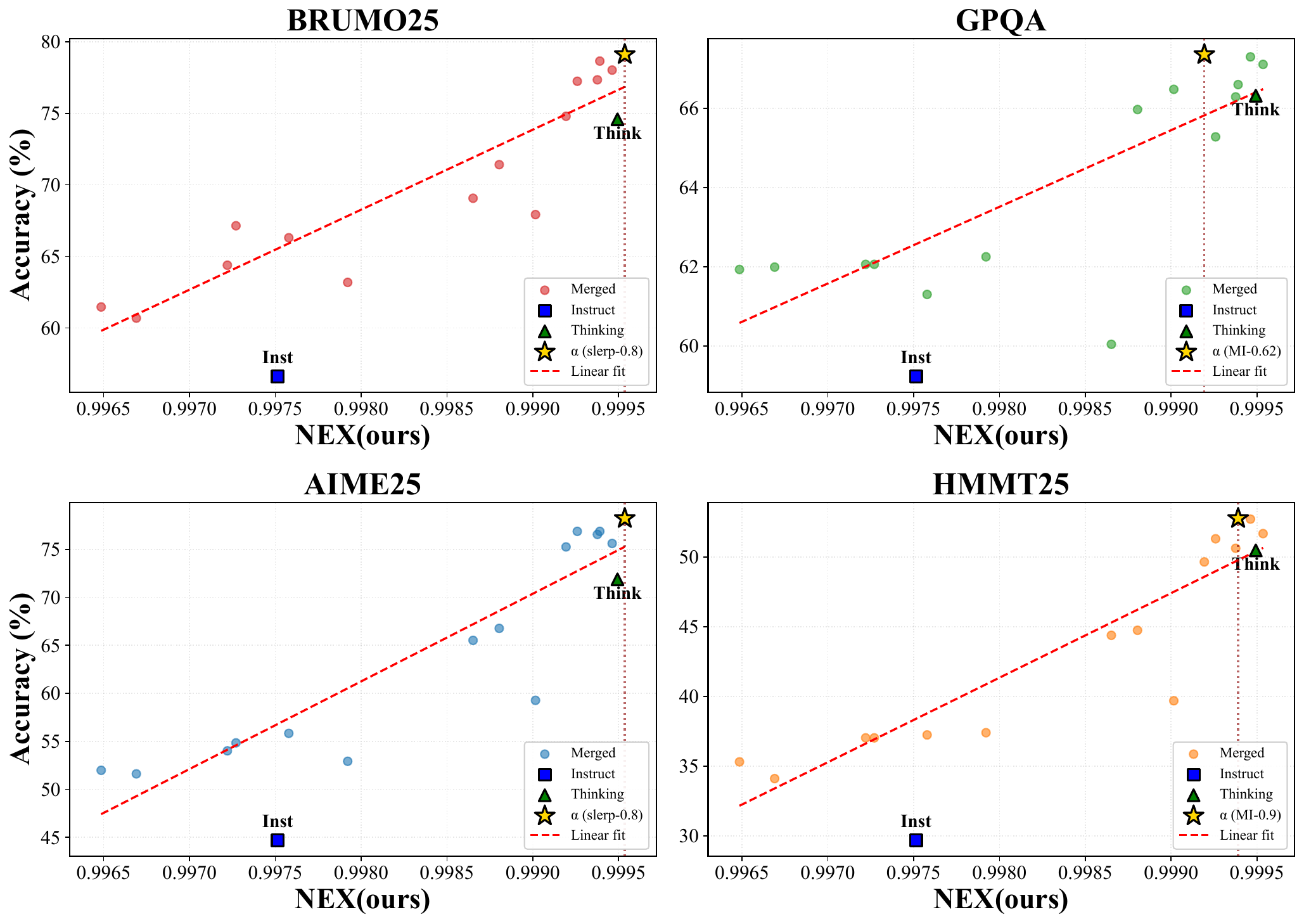}
    \caption{}\label{fig:linear_scaling}
  \end{subfigure}
};

% (c) 右侧图
\node[inner sep=0, right=0.01\textwidth of B] (C) {%
  \begin{subfigure}[b]{0.32\textwidth}
    \centering
      \includegraphics[width=\linewidth]{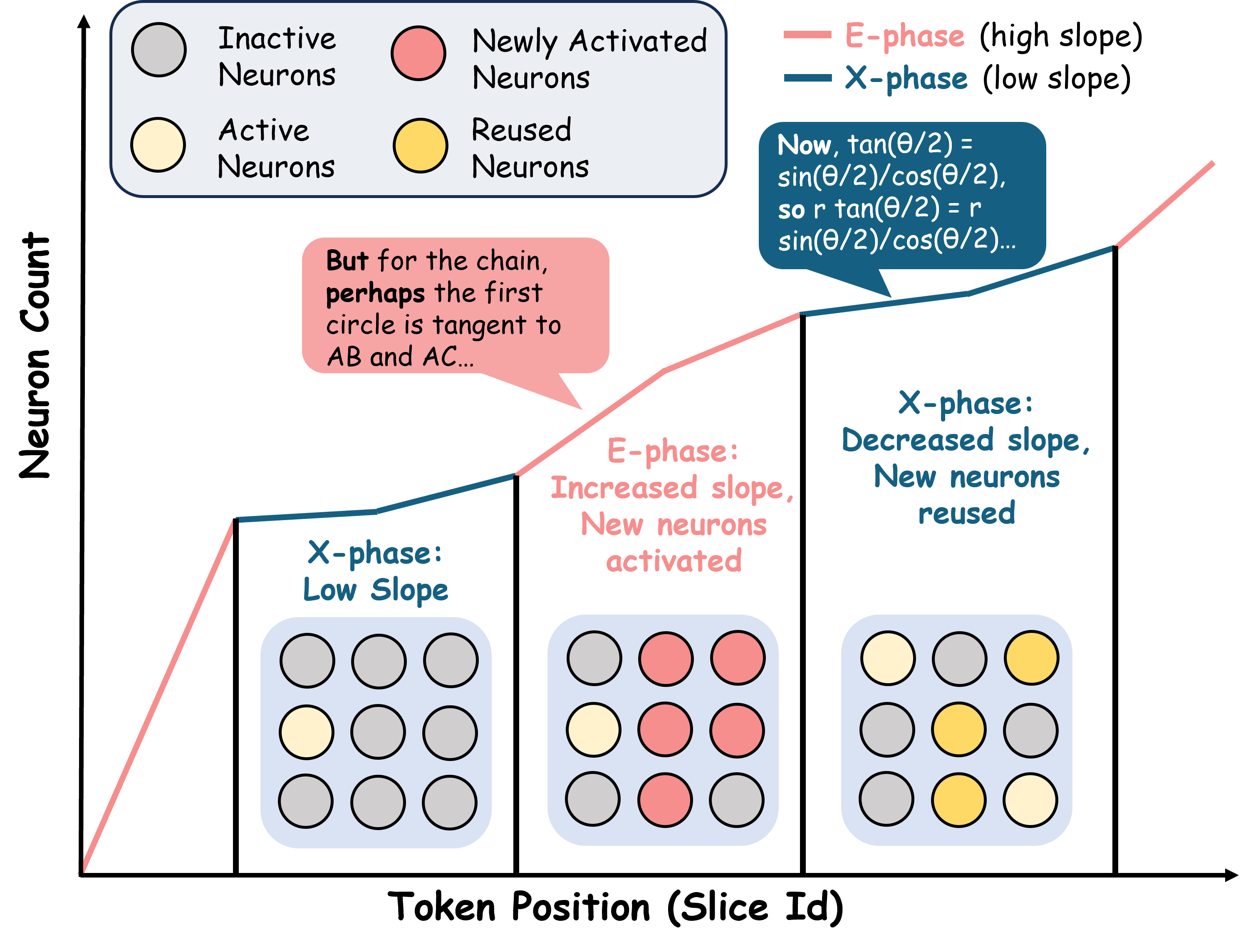}
    \caption{}\label{fig:token_neuron}
  \end{subfigure}
};

% 绘制第一条分割虚线 (A 和 B 之间)，缩短长度
\draw[dashed, line width=0.6pt, gray!50]
  ($(A.north east)!0.5!(B.north west) - (0, 1em)$) --
  ($(A.south east)!0.5!(B.south west) + (0, 1em)$);

% 绘制第二条分割虚线 (B 和 C 之间)，缩短长度
\draw[dashed, line width=0.6pt, gray!50]
  ($(B.north east)!0.5!(C.north west) - (0, 1em)$) --
  ($(B.south east)!0.5!(C.south west) + (0, 1em)$);

\end{tikzpicture}

\caption{\textbf{(a)} Inverted-U relationship between accuracy and an entropy-only exploration proxy (average number of high entropy rows).
Across benchmarks and merged model variants, exploration is beneficial up to a point, after which additional
high entropy reasoning correlates with reduced accuracy.
\textbf{(b)} Strong positive linear correlation between accuracy and the proposed NEX score across four benchmarks. The consistent linear fit demonstrates that unlike raw entropy, NEX provides a monotonic signal where higher scores robustly predict better reasoning performance
\textbf{(c)} Schematic of the token--neuron dynamics curve: high-slope segments (red) correspond to E-phase where the model recruits new neurons to branch into hypotheses, while low-slope segments (blue) correspond to X-phase where the model reuses existing circuits to execute calculations.}
\label{fig:pair}
\end{figure*}

This phenomenon also suggests that an entropy-only view, while useful, is often too coarse to characterize the full
exploration--exploitation dynamics of a CoT.
Token entropy is a natural proxy for predictive uncertainty
\citep{shannon1948communication,kadavath2022lmsknow},
and high entropy tokens can align with decision points where the model must choose among competing continuations
\citep{wang2025beyond8020}.
However, exploration in a CoT is inherently a process that unfolds over multiple steps: the model may open a branch,
accumulate evidence, and then either consolidate and exploit a promising direction or continue to branch.
Consequently, entropy alone provides limited information about how exploration proceeds and, in practice, does not
reliably separate effective high entropy decision points that lead to progress from ineffective high entropy points that
reflect confusion or unproductive wandering \citep{kuhn2023semantic,welleck2024metageneration}.

To address this gap, we turn to internal activations.
Recent evidence suggests that uncertainty signals such as entropy can be viewed as low-dimensional summaries of richer
neuron-level dynamics \citep{cao2025modelutilitylawevaluating,chen2025llmssignaltheyreright}.
We model a CoT trace from a token--neuron temporal perspective: we track the rate at which a trace recruits previously
unused MLP neurons over time, forming a \emph{novelty-slope} time series.
We find that high recruitment slope segments often correspond to exploration, e.g. branching into new hypotheses, while low
slope segments correspond to exploitation, e.g. reusing and consolidating existing circuits to complete the solution.

Based on this perspective, we design \NEX, a label-free method for scoring neuron dynamics.
\NEX segments each CoT into E-phase (exploration) and X-phase (exploitation) by fitting a sticky two-state HMM to a processed novelty-slope (log-transformed, de-trended, and MAD-normalized).
Neurons first introduced during E-phase are credited positively if subsequently reused in X-phase, associated with productive exploration, or negatively otherwise, indicating redundant exploration.
These neuron-level efficiency weights are aggregated into a response score via a simple Good-Mass Fraction: the fraction of activation mass on positively weighted neurons. Our contributions are summarized as follows:
% The resulting score provides a unified signal for three practical tasks:
% selecting the best checkpoint within a model-merge or trained-mixture family;
% choosing among competing CoT traces at inference time;
% and curating high-quality training data for student distillation.
% With \NEX score as a unified signal, we investigate the following research questions: 
% provides a unified signal for three practical tasks:
% selecting the best checkpoint within a model-merge or trained-mixture family;
% choosing among competing CoT traces at inference time;
% and curating high-quality training data for student distillation.
% \subsection{Research questions}

% \begin{itemize}[leftmargin=1.2em]

% \paragraph{Our contributions:}
\begin{itemize}[leftmargin=1.2em]
  \item We propose a token--neuron temporal view of CoT reasoning and use a sticky HMM over novelty-slope dynamics to segment traces into E-phase and X-phase in a label-free and temporally consistent manner.
  \item We introduce \NEX, a label-free neuron-dynamics score that ranks candidate model variants and reasoning traces, and can also be used as a practical signal for filtering training data without access to task labels.
  \item We provide mechanistic validation via cross-model neuron transfer: transplanting neurons associated with productive exploration improves accuracy, whereas transferring neurons associated with redundant exploration degrades accuracy, supporting the causal relevance of the learned neuron weights.
\end{itemize}

% With \NEX score as a unified signal, we investigate the following research questions: 

%   \textbf{RQ1 (Segmentation validity):} Does neuron-based HMM segmentation capture exploration/exploitation phases more reliably than entropy thresholding? We validate via human agreement and comparative analysis with entropy baselines.

%   \textbf{RQ2 (Model selection):} Can \NEX, learned from a small unlabeled mini-set, reliably rank and select models under distribution shift for both merged and trained-mixture models? How does ranking quality scale with mini-set size?

%   \textbf{RQ3 (Monotonicity):} Raw E-phase ratio exhibits an inverted-U relationship with accuracy. Does neuron weighting linearize this relationship, making \NEX score monotonically predictive of correctness?

%   \textbf{RQ4 (Data curation):} Can \NEX serve as a data curation signal that improves student training under equal token budgets, outperforming entropy- and length-based baselines?

%   \textbf{RQ5 (Causal evidence):} Are the learned ``good'' and ``bad'' neurons causally important? What happens when we mask them during inference?
% % \end{itemize}

%% file: sections/related_work.tex
\section{Related Work}

\textbf{Model selection and ranking for checkpoint variants and merges.}
Parameter-space ensembling and model merging generate large families of candidate checkpoints, including model soups and interference-aware merging \citep{wortsman2022soups, yadav2023ties}, as well as task-vector composition and weighted averaging \citep{ilharco2023taskarithmetic, matena2022fishermerging}.
Recent work shows that merge quality depends strongly on the base model, architecture, and fine-tuning conditions,
and provides systematic evaluations and definitions of mergeability for LLMs \citep{hitit2025mergingstudy, rahamim2026mergeability}.
In practice, selecting among merge candidates often relies on behavior-level proxies
(e.g., held-out loss/perplexity) or labeled benchmarks \citep{wortsman2022soups, hitit2025mergingstudy}.
In contrast, \NEX targets \emph{reasoning trace quality} and ranks both merged and trained-mixture variants
using only a small unlabeled activation set, leveraging internal neuron dynamics rather than output statistics.

\textbf{Mechanistic signals from activations.}
Activation-based analyses are widely used for interpretability and diagnostics
(e.g., representation quality across layers) \citep{skean2025layerbylayer}.
Recent work also suggests that internal activation patterns can support practical evaluation and selection,
including mechanism-interpretable evaluation metrics \citep{cao2025modelutilitylawevaluating}
and selection based on neuron agreement decoding \citep{chen2025llmssignaltheyreright}.
Building on these insights, our method operationalizes sparse neuron activation dynamics with a simple temporal model (HMM)
to produce a deployable score that unifies model ranking, trace selection, and score-based data curation.

% \textbf{Data curation and curriculum for LMs.}
% Data filtering and curriculum methods aim to improve training efficiency by selecting higher-quality or appropriately difficult data
% \citep{xie2024doremi,engstrom2024dsdm}.
% Existing approaches typically rely on loss-based or perplexity-based criteria.
% We add a neuron-dynamics-based score designed specifically for \emph{reasoning trace quality} that requires neither correctness labels nor additional models.

%% file: sections/method.tex
\section{Methodology}
\label{sec:method}

% preamble

% in body (put it early enough before you discuss it)
\begin{figure*}[t]
  \centering
  \includegraphics[width=\textwidth]{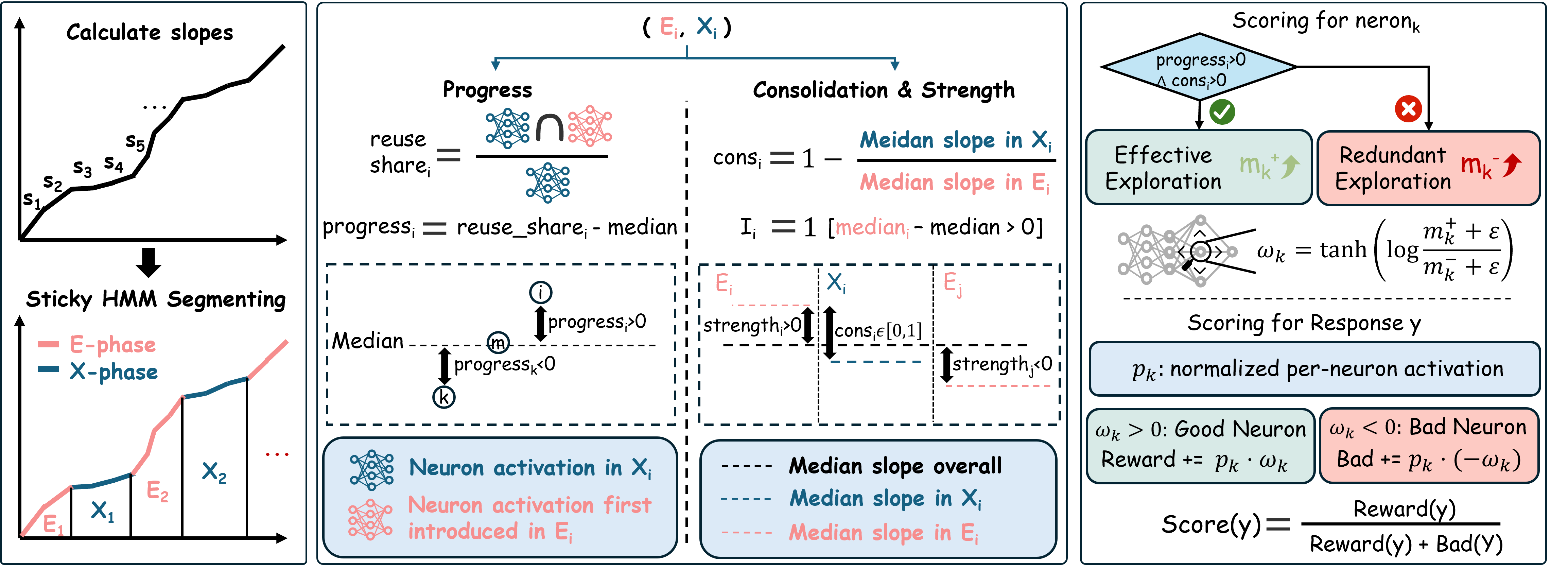}
  \caption{Overview of the NEX algorithm.  \textbf{Left}: Calculation of novelty-slope time series and E-X segmentation via sticky HMM.
  \textbf{Middle}:
  Progress via new-neuron reuse, consolidation (slope drop) and strength gating during E$\rightarrow$X cycles.
  \textbf{Right}:
  Scoring the neurons with normalized weights and the responses via neuron-weighted activation mass.
  }
  \label{fig:main}
\end{figure*}

% Given a CoT response $y=(y_1,\dots,y_T)$, we assign a label-free score $\mathrm{Score}(y)\in[0,1]$ predictive of correctness.
% \NEX uses only within-trace neuron reuse: productive reasoning tends to introduce features during exploration and reuse a subset during consolidation.
% All components are unsupervised (no answer labels, no benchmark-specific tuning).

We propose \NEX, a label-free scoring method for ranking model variants and reasoning traces.
\NEX proceeds in three steps.
\textbf{In \secref{sec:seg_cycles},} we compute a novelty-slope time series from sparse MLP activations and segment each CoT into
\textbf{E-phase} and \textbf{X-phase} with a sticky 2-state HMM, yielding contiguous
E$\rightarrow$X cycles.
\textbf{In \secref{sec:progress},} for each cycle, we quantify progress and consolidation from neuron reuse and apply a strength gate to
filter weak cycles. 
% we then aggregate cycle outcomes into signed neuron weights.
\textbf{In \secref{sec:scoring},} We aggregate cycle outcomes into signed neuron weights, which are then used to compute a Good-Mass Fraction for each individual response. Finally we score a model by averaging
response scores on an unlabeled mini activation set.
Notation is summarized in Appendix~\ref{app:notation}.

\subsection{E-X segmentation and cycles}
\label{sec:seg_cycles}
% We assume cached sparse MLP activations $a_{k,t}\ge 0$ (top-$\kappa$ per token, Appendix~\ref{app:sparsity}), where neuron key
% $k=(\mathrm{layer}\ll 16)\,|\,\mathrm{unit}$.
% We split the response into rows of $W$ tokens (we use $W{=}32$, Appendix~\ref{app:rows}) and let $T_r$ be token indices in row $r$.
% Define activated neurons in row $r$ as $\mathcal{N}_r=\{k:\exists t\in T_r,\,a_{k,t}>\tau\}$ (we use $\tau{=}0$),
% and $\mathcal{N}_{<r}=\cup_{j<r}\mathcal{N}_j$.
% The novelty slope is
% \begin{equation}
% s_r=\frac{|\mathcal{N}_r\setminus \mathcal{N}_{<r}|}{|T_r|}.
% \label{eq:slope}
% \end{equation}
\paragraph{Novelty slope from sparse MLP activations. }
To analyze internal feature dynamics, we segment the model's response into fixed-size rows, each containing 32 tokens, and identify the set of active sparse MLP neurons $\mathcal{N}_r$ for each row $r$ (see Appendix~\ref{app:sparsity} for logging details). We track the recruitment of new features using the \emph{novelty slope} $s_r$, defined as the token-normalized count of neurons that appear in row $r$ but have not appeared in the preceding history $\mathcal{N}_{<r} = \bigcup_{j<r}\mathcal{N}_j$:
\begin{equation}
    s_r = \frac{|\mathcal{N}_r \setminus \mathcal{N}_{<r}|}{|T_r|},
    \label{eq:slope}
\end{equation}
where $|T_r|$ denotes the number of tokens in row $r$. The resulting time series $\{s_r\}$ quantifies the rate of feature introduction: a high slope suggests \emph{exploration} (recruiting new concepts), while a low slope indicates \emph{exploitation} (reusing existing features or consolidation).
We robustly normalize $\{s_r\}$ into a 1D observation series $z_r$ via log transform, detrending, and MAD standardization (details in Appendix~\ref{app:preprocessing}).

\paragraph{Segmentation via sticky-HMM.}
We segment rows into E-phase vs. X-phase by a 2-state Gaussian HMM over $z_r$,
initialized by an unsupervised 2-component GMM (fixed seed) and decoded by Viterbi with a fixed sticky transition with a stickiness parameter $\rho$, which discourages rapid state switching, encoding the prior that E- and X-phases span multiple rows (more details in Appendix~\ref{app:hmm-details}).

We label the state with larger emission mean as E, extract contiguous E$\rightarrow$X cycles $\{(E_i, X_i)\}_{i=1}^M$, and optionally merge segments shorter than a minimum run length (min run = 2 rows).

\subsection{Progress and consolidation from neuron reuse}
\label{sec:progress}

\paragraph{Cycle-level credit assignment (intuition first).}
% We treat each E$\rightarrow$X cycle $i$ as a self-supervised ``did exploration pay off?'' event.
% During $E_i$ the model introduces new internal features (new neurons). During $X_i$ the model either
% reuses those features to consolidate a direction (productive exploration) or fails to reuse them (redundant exploration).
% \NEX quantifies this per cycle using (i) \emph{progress} (reuse above the run median), (ii) \emph{consolidation}
% (a slope drop from $E_i$ to $X_i$), and (iii) a binary \emph{strength gate} that ignores weak exploration cycles.
% Crucially, we only consider neurons first introduced during $E_i$, so generic always-on circuitry does not dominate.

We treat each E$\rightarrow$X cycle $i$ as a self-supervised "did exploration pay off?" event, where during $E_i$ new neurons are introduced, and during $X_i$ they are either reused for productive exploration or not, representing redundant exploration. \NEX quantifies this using (i) \emph{progress} (reuse above the run median), (ii) \emph{consolidation} (slope drop from $E_i$ to $X_i$), and (iii) a binary \emph{strength gate} to ignore weak cycles, considering only neurons first introduced during $E_i$.We operationalize this cycle-level view in the following steps.

\paragraph{Step 1: identify newly introduced neurons for the cycle.}
New neurons introduced in cycle $i$:
\begin{equation}
N_i = \{k : \exists\,r \in E_i \text{ s.t.\ } k \in \mathcal{N}_r \text{ and } k \notin \mathcal{N}_{<r}\}.
\end{equation}

\paragraph{Step 2: progress via reuse in the subsequent X-phase.}

Let $\mathbf{u}_{\ell,t}\in\mathbb{R}^{d_{\mathrm{ff}}}$ and $\mathbf{g}_{\ell,t}\in\mathbb{R}^{d_{\mathrm{ff}}}$ denote the
pre-activation FFN "up" and "gate" vectors at layer $\ell$ and token position $t$.
We define the gated intermediate vector
\begin{equation}
\mathbf{h}_{\ell,t} \;\triangleq\; \phi(\mathbf{g}_{\ell,t}) \odot \mathbf{u}_{\ell,t},
\end{equation}
where $\odot$ is elementwise multiplication and $\phi(\cdot)$ is the gating nonlinearity.

Each neuron is indexed by a key $k=(\ell,j)$ (layer $\ell$, unit $j$). The logged activation mass for neuron $k$ at token $t$ is
\begin{equation}
a_{k,t} \;\triangleq\; \max\!\bigl(0,\; (\mathbf{h}_{\ell,t})_j \bigr).
\end{equation}

Let $A_{k,r}=\sum_{t\in T_r} a_{k,t}$ be the activation mass of neuron $k$ in row $r$.
We define the \emph{reuse share} of newly introduced neurons in the subsequent exploitation segment:
\begin{equation}
\mathrm{reuse\_share}_i
\;=\;
\frac{\sum_{r\in X_i}\sum_{k\in N_i} A_{k,r}}{\sum_{r\in X_i}\sum_{k} A_{k,r}+\epsilon}.
\label{eq:reuse-share}
\end{equation}

\paragraph{Step 3: within-run centering.}
To avoid introducing additional thresholds and to make cycles comparable within a trace, we center within each run:
\begin{equation}
\mathrm{progress}_i \;=\; \mathrm{reuse\_share}_i - \mathrm{median}\{\mathrm{reuse\_share}_j\}_{j=1}^{M}.
\label{eq:reuse-progress}
\end{equation}
Positive progress indicates the E-phase-introduced neurons are reused more than typical in the subsequent X-phase, suggesting effective exploration.

\paragraph{Step 4: consolidation and strength gating.}
Consolidation requires a real E$\rightarrow$X transition via a slope drop:
\begin{equation}
\mathrm{cons}_i
\;=\;
\mathrm{clip}\!\left(1-\frac{\mathrm{median}\{s_r:r\in X_i\}}{\mathrm{median}\{s_r:r\in E_i\}+\epsilon},\;0,\;1\right).
\label{eq:reuse-cons}
\end{equation}
Strength gating: we gate credit assignment to cycles whose E-phase intensity is above the trace median:
\begin{equation}
\begin{aligned}
\mathrm{strength}_i
&=\mathrm{median}\{s_r:r\in E_i\}-\mathrm{median}\{s_r\}_{r=1}^{R},\\
I_i &= \mathbf{1}[\mathrm{strength}_i>0].
\end{aligned}
\label{eq:reuse-strength}
\end{equation}
Only cycles with $I_i=1$ contribute; \emph{strength} is used as a binary gate, not as a continuous multiplier.
We call an E$\rightarrow$X cycle \emph{effective} if $\mathrm{progress}_i>0$ and $\mathrm{cons}_i>0$; otherwise it is treated as \emph{redundant}.

\subsection{Neuron weights and scoring}
\label{sec:scoring}

\paragraph{From cycle outcomes to neuron weights.}
Across the mini-set, each cycle contributes evidence about the neurons it \emph{introduced}.
We maintain two nonnegative accumulators per neuron: $m_k^+$ for effective exploration mass and $m_k^-$
for redundant exploration mass. Let $\alpha_{k,i}$ be the activation mass of neuron $k$ at its introduction
within $E_i$ (i.e., summing $A_{k,r}$ over rows $r \in E_i$ where $k$ is first-seen in the run).
We then update:
\begin{align}
m_k^{+} &\mathrel{+}= \alpha_{k,i}\cdot I_i\cdot \mathrm{progress}_i\cdot \mathrm{cons}_i,
&&\text{if effective,}\\
m_k^{-} &\mathrel{+}= \alpha_{k,i}\cdot I_i\cdot |\mathrm{progress}_i|,
&&\text{otherwise.}
\end{align}
Finally, we compute a signed neuron weight:
\begin{equation}
w_k \;=\; \tanh\!\Big(\log\frac{m_k^{+}+\epsilon}{m_k^{-}+\epsilon}\Big) \in [-1,1].
\label{eq:weight}
\end{equation}
We refer to neurons with large positive weights as \textbf{effective neurons} and large negative weights as \textbf{redundant neurons}.

\paragraph{Response and model scoring.}

For a new response $y$, let $b_k(y)=\sum_{t=1}^{T} a_{k,t}$ denote the nonnegative activation mass of neuron $k$ accumulated over the response.
Scoring uses only this sparse response-level summary and the learned neuron weights $w_k$; it does \emph{not} re-run the HMM at test time.

\paragraph{Good-Mass Fraction (code-default score).}
Define the positive (good) and absolute (credible) weighted masses:
\begin{align}
\mathrm{PosMass}(y) &= \sum_k b_k(y)\,[w_k]_+, \\
\mathrm{AbsMass}(y) &= \sum_k b_k(y)\,|w_k|.
\end{align}
We score a response by the fraction of credible activation mass that lies on positively weighted neurons:
\begin{equation}
\mathrm{Score}(y)=
\begin{cases}
\dfrac{\mathrm{PosMass}(y)}{\mathrm{AbsMass}(y)}, & \mathrm{AbsMass}(y)>0,\\
0, & \text{otherwise.}
\end{cases}
\label{eq:final-score}
\end{equation}

\paragraph{Interpretable auxiliaries (optional, for logging).}
Let $\mathrm{TotMass}(y)=\sum_k b_k(y)$. We may also report
\begin{align}
\mathrm{Reward}(y) &= \mathrm{PosMass}(y)/\mathrm{TotMass}(y), \\
\mathrm{Bad}(y) &= (\mathrm{AbsMass}(y)-\mathrm{PosMass}(y))/\mathrm{TotMass}(y),
\end{align}
in which case $\mathrm{Score}(y)=\mathrm{Reward}(y)/(\mathrm{Reward}(y)+\mathrm{Bad}(y))$.

\paragraph{Model score:} average $\mathrm{Score}(y)$ over prompts in the unlabeled mini-set.

\paragraph{Data score:} apply the same response score to teacher-generated training samples to filter data.

% \medskip
% Algorithm~\ref{alg:reusehmm} summarizes the full procedure (See Appendix~\ref{app:code}).

%% file: sections/experimental_setup.tex
\section{Experimental Setup}
\label{sec:setup}

\subsection{Candidate model families}
\label{sec:model-families}

We evaluate \NEX on families of closely related model variants that span the E-X spectrum.
Each family is anchored by two endpoint models with distinct reasoning behaviors:
\begin{itemize}[leftmargin=1.2em]
  \item \textbf{Instruct endpoint:} Instruction-tuned for direct answers, typically under-exploring the feature space.
  \item \textbf{Thinking endpoint:} Tuned for deliberation, typically over-exploring via verbose reasoning.
\end{itemize}

\paragraph{Merged model families.}
We generate intermediate model variants via parameter-space merging between the two endpoints using MI-$\lambda$ and SLERP-$t$, yielding ${\sim}20$ candidates per family.
Among publicly available Qwen3 text-only models, Qwen3-4B is the only small-scale model that offers separate Instruct and Thinking variants.
To study scaling behavior, we additionally use the Qwen3-VL series (4B, 8B, 32B), which offers both endpoints across multiple sizes.
Our main model-ranking results are thus reported on four merged families: Qwen3-4B, Qwen3-VL-4B, Qwen3-VL-8B, and Qwen3-VL-32B (full model list in Appendix~\ref{app:model-list}).

\paragraph{Trained model families.}
We also evaluate on model variants obtained by \emph{real training} on mixed Instruct/Thinking rollout data.
Specifically, we use Qwen3-30B-A3B-Instruct and Qwen3-30B-A3B-Thinking to generate teacher rollouts on three public math reasoning datasets (\texttt{dapo-math-17k}, \texttt{deepmath-103k}, \texttt{deepscaler}), then train student models on different mixtures of Instruct vs.\ Thinking traces (details in Appendix~\ref{app:rollout-data}).

Unless stated otherwise, we use the Good-Mass Fraction scoring variant (Eq.~\ref{eq:final-score}).

\subsection{Data sources}
\label{sec:data}

\paragraph{NEX weight learning on mini activation set.}
We compute all self-supervised quantities including HMM segmentation, progress, consolidation, and neuron weights on a small unlabeled mini set of $N=100$ problems.
The mini set is drawn from the same rollout pool as the trained model families but is held out from training, ensuring no overlap; no ground-truth answers are used for scoring.

\paragraph{Downstream accuracy on evaluation benchmarks.}
We evaluate model ranking on five hard reasoning benchmarks:
AIME24~\citep{aime24},
AIME25~\citep{aime25},
GPQA~\citep{rein2023gpqa},
HMMT25~\citep{hmmt25},
and BRUMO25~\citep{brumo25},
using exact-match accuracy.

\paragraph{Data curation on teacher rollout pool (RQ4).}
For score-based data curation experiments (\secref{sec:rq4}), we score the full rollout pool with the exception of the mini-set, filtering training data based on their \NEX scores; full details are provided in Appendix~\ref{app:rollout-data}.

\subsection{Decoding settings}
\label{sec:generation}

All experiments, including model ranking, trace-level selection, and training data generation, use a unified sampling setup with temperature $T{=}0.7$, top-$p{=}0.9$, and a maximum of $32{,}768$ generated tokens.

\subsection{Baselines}
\label{sec:baselines}

We compare \NEX to simple label-free baselines computed from the same candidate traces: \textbf{(1) Length}, number of tokens in response . (2) \textbf{High-Entropy Sum (HES)}, sum token entropy from the next-token distribution~\citep{anonymous2026unified}.\textbf{(3) Top-20\% entropy fraction}, fraction of rows whose entropy falls in the top 20\% within the response. \textbf{(4) Log-prob}, mean token log-probability under the generating model.

% \paragraph{Segmentation baselines (RQ1).}
% For RQ1, we additionally compare our HMM-based segmentation against threshold-based alternatives:
% (1) \emph{entropy thresholding}: label a row as E-phase if its mean token entropy exceeds the trace median;
% (2) \emph{novelty-slope thresholding}: label a row as E-phase if its novelty slope exceeds the trace median.
% These baselines lack the temporal smoothing of the HMM.

% \subsection{Metrics}
% \label{sec:metrics}

% \paragraph{Segmentation validation.}
% Block-level agreement with human E/X labels, reported as accuracy and Cohen's $\kappa$.

% \paragraph{Checkpoint ranking.}
% Pearson correlation between the mean mini-set score and downstream accuracy across candidate checkpoints, along with
% Regret@1 (percentage points) and Hit@3.

% \paragraph{Neuron transfer.}
% $\Delta$Accuracy after transplanting neurons with positive vs.\ negative \NEX weights across models.

% \subsection{Implementation details}
% \label{sec:impl}
% Implementation details (software, preprocessing, hyperparameters, and seeds) are provided in Appendix~\ref{app:impl}.

% \paragraph{Segmentation baselines (RQ1).}
% For RQ1, we additionally compare our HMM-based segmentation against threshold-based alternatives:
% (1) \emph{entropy thresholding}: label a row as E-phase if its mean token entropy exceeds the trace median;
% (2) \emph{novelty-slope thresholding}: label a row as E-phase if its novelty slope exceeds the trace median.
% These baselines lack the temporal smoothing of the HMM.

\subsection{Implementation details}
\label{sec:impl}

HMM segmentation uses Python package \texttt{hmmlearn}; student model training uses LLaMA-Factory on 8$\times$H20 GPUs with DeepSpeed ZeRO.
Training hyperparameters: Qwen3-4B-Base, full fine-tuning, batch size 8, learning rate $10^{-5}$, 1 epoch, cosine scheduler with 10\% warmup.
Additional details are provided in Appendix~\ref{app:impl}.

%% file: sections/results_fzk.tex
\section{Results}
\label{sec:results}

% ===================================================================
% RQ1: Neuron-based HMM vs Entropy Thresholding
% ===================================================================
\subsection{RQ1: Neuron-based HMM vs entropy thresholding for E/X segmentation}
\label{sec:rq1}

We validate that our neuron-based HMM segmentation captures E- and X-phases reliably.
We present three layers of evidence: (1) human agreement with HMM labels, (2) novelty slope captures information beyond entropy, and (3) E/X ratios vary systematically across model families.

\paragraph{Human validation.}
We sample problems from the mini activation set and label all HMM-segmented blocks, yielding $2{,}858$ labeled blocks.
Two independent annotators label each block as E- or X-phase based on textual markers (see Appendix~\ref{app:human} for annotation guidelines and example cases).

\begin{table}[t]
\centering
\small
\caption{Human agreement with HMM-based E-X segmentation on the mini activation set ($100$ problems from Qwen3-4B-Thinking).}
\label{tab:human}
\begin{tabular}{lccc}
\toprule
Type & \#Blocks & Correct & Acc. \\
\midrule
Overall & 2,858 & 2,562 & 89.6\% \\
E-phase & 1,429 & 1,404 & 98.2\% \\
X-phase & 1,429 & 1,158 & 81.0\% \\
\bottomrule
\end{tabular}
\end{table}

\tabref{tab:human} shows strong agreement between HMM segmentation and human labels, with higher accuracy on E-phase than X-phase.
The dominant failure mode is over-detection of E-phase: the HMM occasionally labels exploitation blocks as exploration.

\paragraph{Novelty slope captures information beyond entropy.}
Token entropy and novelty slope show moderate correlation ($\rho = 0.43$), with $34.7\%$ of rows in disagreement quadrants.
Manual inspection shows novelty slope more accurately reflects exploration: rows where high novelty coincides with low entropy often contain structured hypothesis generation that entropy misses (see Appendix~\ref{app:novelty-entropy}).

\begin{figure}[t]
  \centering
  \includegraphics[width=0.75\linewidth]{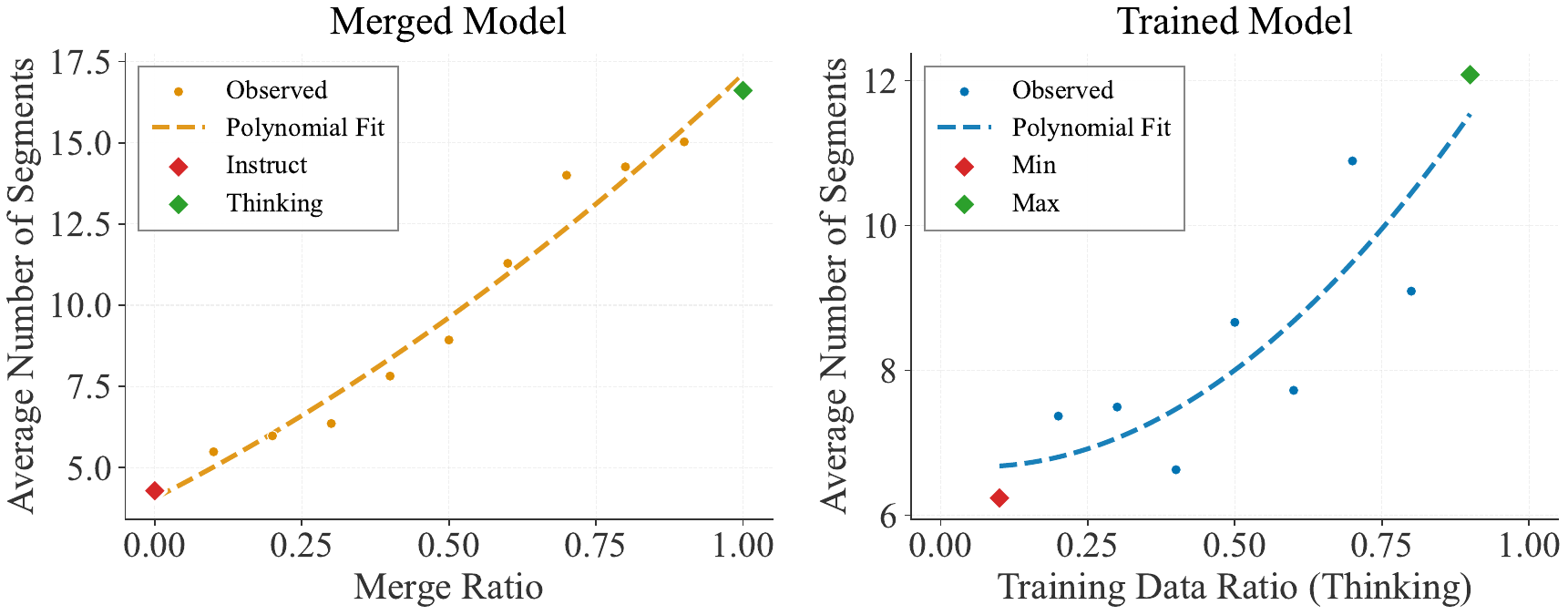}
  \caption{Average number of E-phase segments across model families. \textbf{Left:} Merged Qwen3-4B models with SLERP between Instruct and Thinking show smooth quadratic growth. \textbf{Right:} Trained Qwen3-4B-Base models by varying Instruct-Thinking data mixtures exhibit moderate growth with higher variance.}
  \label{fig:ex_ratio}
\end{figure}

\paragraph{E/X ratios differ across model families.}
To validate that HMM segmentation captures meaningful behavioral differences, we compare E-phase segment counts across two model construction methods (\secref{sec:model-families}).
As shown in \figref{fig:ex_ratio}, merged Qwen3-4B models (SLERP interpolation) exhibit smooth growth in segment count as the merge ratio shifts toward Thinking, while trained Qwen3-4B-Base models show more moderate growth with higher variance.
This contrast confirms that merged models inherit exploration patterns continuously from endpoints, while trained models develop distinct dynamics through gradient-based optimization.

% ===================================================================
% RQ2: From Inverted-U to Monotonicity (Mechanism Explanation)
% ===================================================================
\subsection{RQ2: From inverted-U to monotonicity via neuron weighting}
\label{sec:rq2}

We show that raw exploration level is non-monotonic with accuracy, and that neuron weighting converts it into a monotonic selection signal.
We use 19 Qwen3-4B models spanning from Instruct, Thinking baselines to 17 merged variants via MI and SLERP, and sample 3 generations per prompt on the 100-problem mini set.

\begin{figure}[t]
  \centering
  \includegraphics[width=0.75\linewidth]{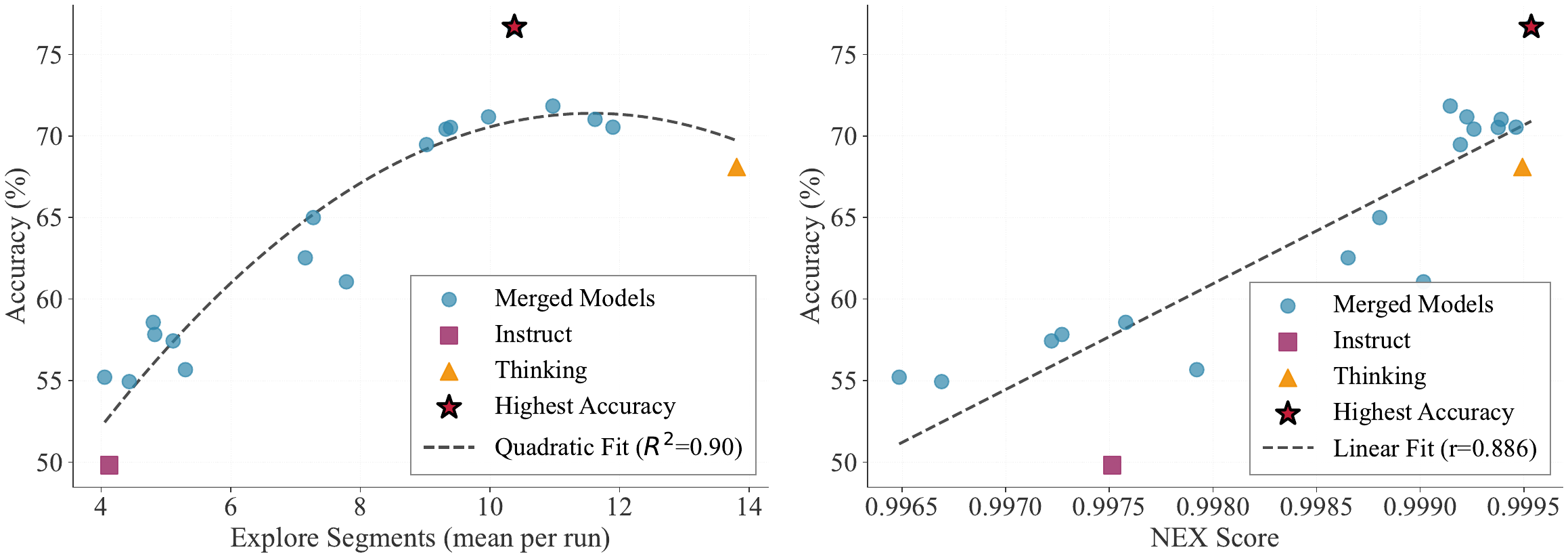}
  \caption{Exploration-accuracy relationship across 19 Qwen3-4B models.
  \textbf{Left:} Mean E-phase segments vs.\ accuracy exhibits an inverted-U ($R^2 = 0.90$), with optimal performance at ${\sim}11.6$ segments.
  \textbf{Right:} \NEX score vs.\ accuracy shows strong positive correlation ($r = 0.886$), demonstrating that neuron weighting linearizes the relationship.
   Markers: merged models (blue circles), Instruct baseline (magenta square), Thinking baseline (orange triangle), best model (star).}
  \label{fig:inverted_u}
\end{figure}

\paragraph{Raw exploration level is non-monotonic.}
We quantify exploration by the mean number of E-phase segments per run.
As shown in \figref{fig:inverted_u} (left), this raw count exhibits a significant inverted-U relationship with benchmark accuracy ($R^2 = 0.90$), with peak performance at approximately 11.6 segments.
Both under-exploration and over-exploration correlate with lower accuracy, making the raw count unsuitable as a direct ranking signal (see Appendix~\ref{app:inverted-u-extended} for extended analysis across model families).

\paragraph{Neuron weighting linearizes the relationship.}
The inverted-U arises because raw exploration counts conflate \emph{productive} exploration (neurons reused in X-phase) with \emph{redundant} exploration (neurons soon discarded).
\NEX addresses this by assigning $w_k > 0$ to neurons in productive E$\rightarrow$X cycles and $w_k < 0$ to those in redundant cycles.
As shown in \figref{fig:inverted_u} (right), the \NEX score exhibits strong positive correlation with accuracy ($r = 0.864$), confirming that efficient neuron reuse is a key indicator of model performance.
Appendix~\ref{app:effective-redundant} illustrates representative cases where both types of exploration occur within the same reasoning trace.
We validate this distinction causally in \secref{sec:rq5}: transplanting $w_k > 0$ neurons from Thinking to Instruct models improves accuracy, while transplanting $w_k < 0$ neurons degrades it.

% ===================================================================
% RQ3: Label-Free Model Ranking (Application)
% ===================================================================
\subsection{RQ3: Label-free model ranking}
\label{sec:rq3}

\paragraph{Main results.}
We evaluate whether \NEX can rank candidate model variants without access to answers under distribution shift.
For each candidate model variant, we compute mean \NEX score on the same  unlabeled mini activation set and compare it with held-out benchmark accuracies.
We use a \emph{self-calibrated} setting where each candidate trains its own neuron weights from its own mini-cache.
\tabref{tab:corr} summarizes performance by series.
\begin{wrapfigure}[18]{r}{0.5\linewidth}
  \centering
  \includegraphics[width=\linewidth]{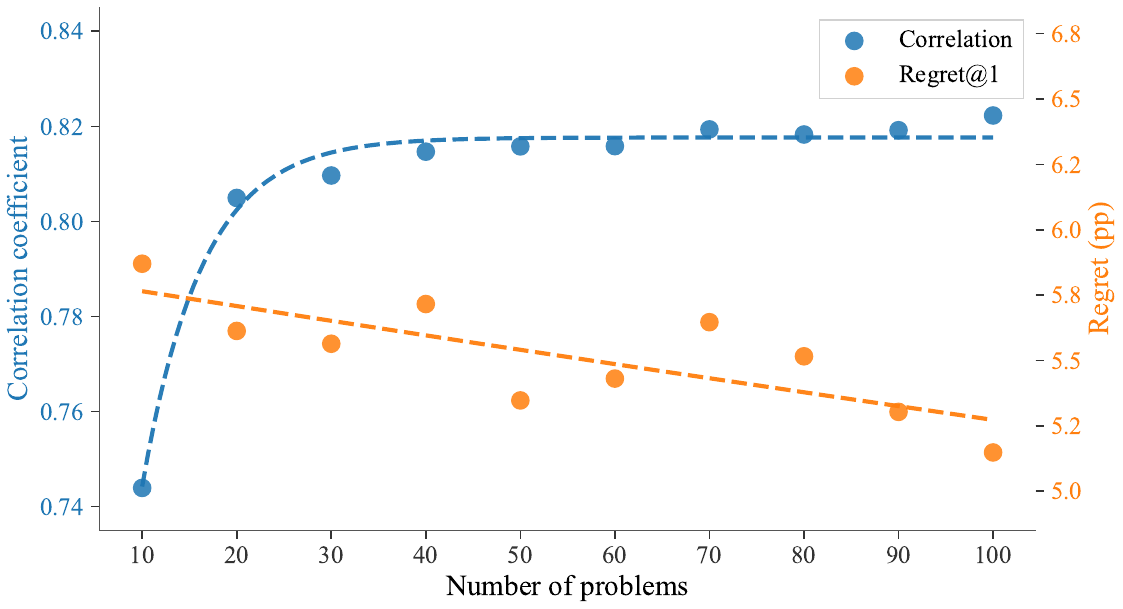}
  \caption{Sample efficiency of \NEX for model selection. We train \NEX weights on $N$ problems and compute the correlation between \NEX scores and accuracies across all merged models. Correlation (blue, left) saturates rapidly; Regret@1 (orange, right) declines steadily. Scatter: means over 10 seeds; curves: exponential saturation fits.}
  \label{fig:mini_set_size}
\end{wrapfigure}
Across 20 (series, benchmark) pairs spanning four Qwen3 families,
we obtain mean Pearson correlation $r=0.778$, average regret@1 of $2.67$ percentage points, and Hit@3 of $7/20$ (35.0\%).

\begin{figure}[t]
  \centering
  \includegraphics[width=0.8\linewidth]{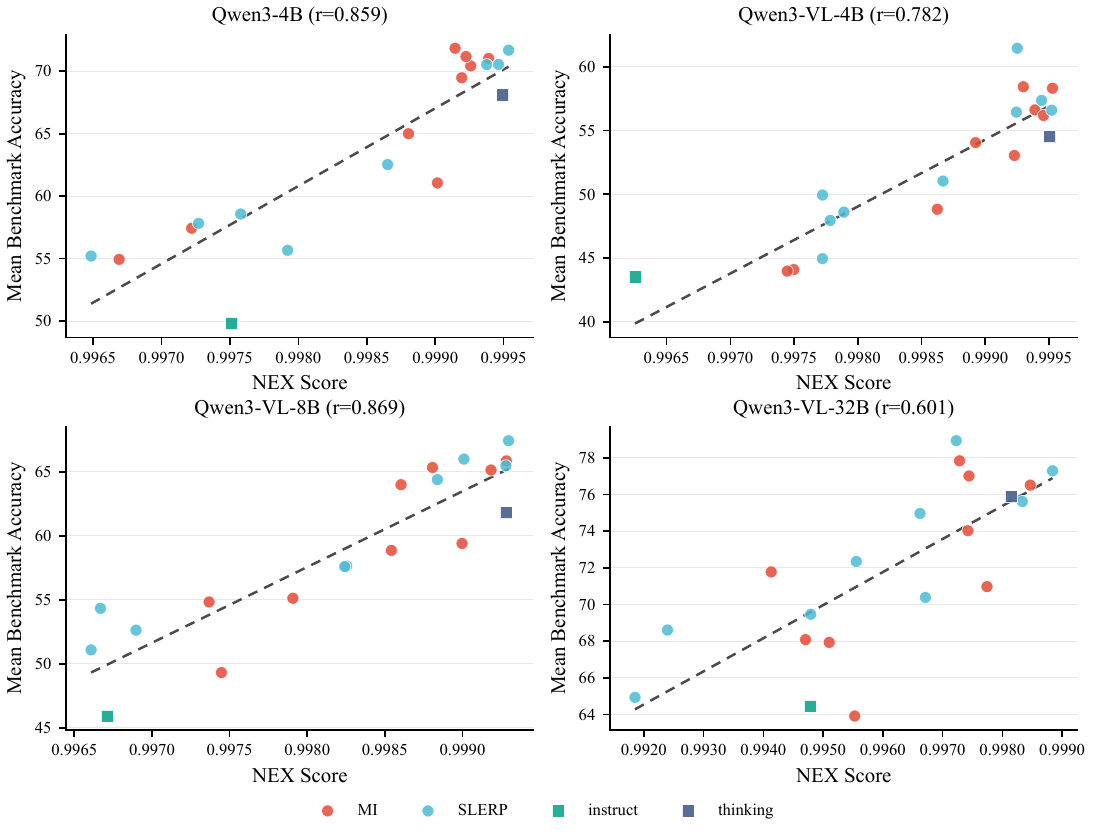}
  \caption{
  Cross-dataset generalization: mean \NEX score on the mini activation set vs.\ test accuracy
  for multiple benchmarks and many merged model variants.
  The Pearson correlation coefficients ($r$) are annotated in each subplot.
  }
  \label{fig:cross_dataset}
\end{figure}

\begin{table}[t]
\centering
\caption{Cross-dataset model selection using \NEX scores on the 100-problem mini set.
Higher is better for $r$ and Hit@3; lower for Regret.}
\label{tab:corr}
\small
\begin{tabular}{@{}lcccc@{}}
\toprule
Series & \#Bench & Pearson $r$ & Regret@1 & Hit@3 \\
\midrule
Qwen3-4B      & 5  & 0.859 & 0.90 & 1/5 \\
Qwen3-VL-4B   & 5  & 0.782 & 5.70 & 2/5 \\
Qwen3-VL-8B   & 5  & 0.869 & 2.04 & 4/5 \\
Qwen3-VL-32B  & 5  & 0.601 & 2.05 & 0/5 \\
\midrule
Overall       & 20 & 0.778 & 2.67 & 7/20 \\
\bottomrule
\end{tabular}
\end{table}

\paragraph{Comparison to baselines.}

\tabref{tab:ranking_baselines} reports the overall model-selection performance of \NEX on the 20 (series, benchmark) pairs in our evaluation.
Compared to simple heuristics including length, entropy, and mean log-probability, \NEX achieves the strongest correlation with downstream accuracy and substantially improves Hit@3, indicating more reliable top-rank selection.
% \begin{wrapfigure}[22]{r}{0.55\linewidth}
%   \centering
%   \includegraphics[width=\linewidth]{figs/miniset_size_curve.pdf}
%   \caption{Sample efficiency of \NEX for model selection. We train \NEX weights on $N$ problems and compute the correlation between \NEX scores and accuracies across all merged models. Correlation (blue, left) saturates rapidly; Regret@1 (orange, right) declines steadily. Scatter: means over 10 seeds; curves: exponential saturation fits.}
%   \label{fig:mini_set_size}
% \end{wrapfigure}
\begin{table}[t]
\centering
\caption{Average cross-dataset model selection performance (20 series$\times$benchmark pairs).}
\label{tab:ranking_baselines}
\small
\begin{tabular}{lccc}
\toprule
Method & Pearson $r$ & Regret@1 (pp) & Hit@3 \\
\midrule
Length        & 0.743 & 6.22 & 0.100 \\
HES & 0.748 & 6.22 & 0.100 \\
Log-prob      & 0.074 & 8.96 & 0.000 \\
\rowcolor{gray!15}
\NEX (ours)   & \textbf{0.778} & \textbf{2.67} & \textbf{0.350} \\
\bottomrule
\end{tabular}
\end{table}

\paragraph{Ablation: which neurons to credit.}
We ablate the choice of which neurons receive credit during E$\rightarrow$X cycles.
Compared to our default approach of crediting only neurons first introduced during E-phase,
attributing cycle credit to \emph{all active neurons} degrades ranking correlation
(Pearson $r$: $0.778 \to 0.673$) and increases selection regret ($2.67 \to 3.83$\,pp).
This confirms that only neurons newly recruited during E-phase carry the productive-vs-redundant distinction.

% \begin{wrapfigure}[22]{r}{0.55\linewidth}
%   \centering
%   \includegraphics[width=\linewidth]{figs/miniset_size_curve.pdf}
%   \caption{Sample efficiency of \NEX for model selection. We train \NEX weights on $N$ problems and compute the correlation between \NEX scores and accuracies across all merged models. Correlation (blue, left) saturates rapidly; Regret@1 (orange, right) declines steadily. Scatter: means over 10 seeds; curves: exponential saturation fits.}
%   \label{fig:mini_set_size}
% \end{wrapfigure}
\paragraph{Mini-set size.}
We vary the mini activation set size $N \in \{10,20,\dots,100\}$ to quantify how many problems are needed for reliable model selection.
For each $N$, we train \NEX weights on $N$ problems and evaluate the correlation between \NEX scores and accuracies across all merged models.
As shown in \figref{fig:mini_set_size}, correlation saturates rapidly and $40$--$60$ problems suffice for near-optimal model selection.

% ===================================================================
% RQ4: Score-Based Data Curation
% ===================================================================
\subsection{RQ4: Score-based data curation improves student training}
\label{sec:rq4}

We demonstrate that \NEX can serve as a practical data curation signal through a three-step validation:
(1)~Best-of-$n$ selection shows that \NEX distinguishes per-sample quality;
(2)~human preference annotation validates that higher \NEX scores align with human-preferred reasoning among correct answers;
(3)~training experiments confirm that \NEX-based filtering improves student performance .

\subsubsection{Best-of-$n$ selection}

\begin{wrapfigure}[14]{r}{0.5\linewidth}
  \centering
  \includegraphics[width=\linewidth]{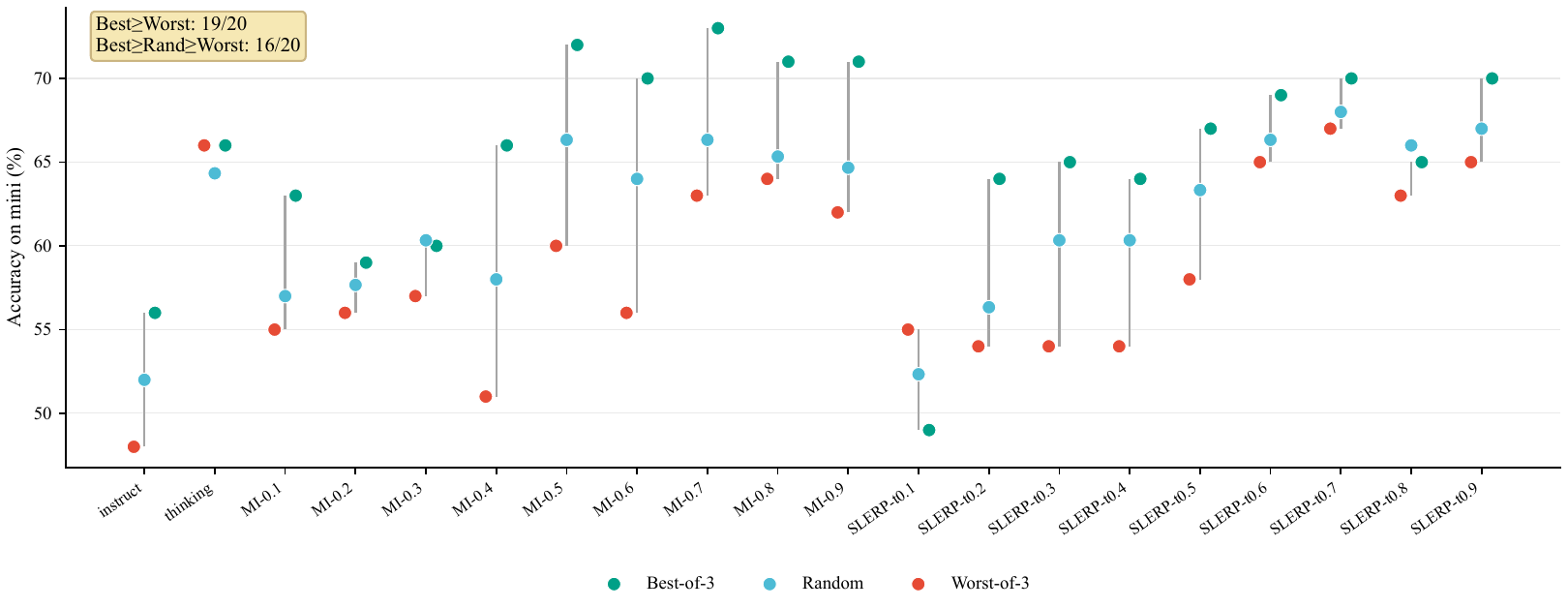}
  \caption{
  Best-of-$n$ selection ($n=3$) on Qwen3-VL-4B models. For each model variant, we sample 3 CoTs per prompt and compare selecting by highest \NEX score (Best-of-3), random selection (Random), and lowest \NEX score (Worst-of-3). Best-of-3 outperforms Worst-of-3 in 19/20 models; the ordering Best $\ge$ Random $\ge$ Worst holds in 16/20 models.
  }
  \label{fig:best_of_n}
\end{wrapfigure}
We first verify that \NEX captures per-sample quality variation.
For each prompt, we sample $n=3$ CoT responses and select the one with the highest or lowest \NEX score.
\figref{fig:best_of_n} shows that selecting the highest-\NEX sample yields consistently higher accuracy than selecting the lowest-\NEX sample, confirming that \NEX distinguishes response quality within the same prompt.

\subsubsection{CoT quality among correct answers}

Beyond accuracy, we verify that \NEX captures reasoning quality differences among correct answers.
Among Best-of-$n$ trials where multiple samples produce correct answers, \NEX-selected CoTs consistently exhibit fewer backtracking markers, shorter token lengths, and clearer structure compared to randomly sampled CoTs (Appendix~\ref{app:best-of-n-cot}).

\subsubsection{Data curation for student training}

Finally, we test whether \NEX-based filtering improves downstream student training.

\paragraph{Protocol.}
We use a teacher model to generate a large pool $\mathcal{D}$ of training traces.
We compute \NEX score and HES score for each sample and construct equal-budget training sets:
\textbf{\NEX-Top20\%} and \textbf{HES-Top20\%}.
We train the same student architecture from the same initialization under identical hyperparameters. As shown in table~\ref{tab:data_curation}, training outcomes for two curation methods reveal that NEX-Top20\% achieves higher scores compared to HES-Top20\% across various tasks, demonstrating the effectiveness of NEX.

\begin{table}[t]
\centering
\caption{Student training outcomes under equal token budgets.}
\label{tab:data_curation}
\footnotesize
\setlength{\tabcolsep}{3pt}
\begin{tabular}{@{}lccccc@{}}
\toprule
Curation & AIME24 & AIME25 & GPQA & BRU25 & HMM25 \\
\midrule
\NEX-Top20 & \textbf{13.3} & \textbf{20.0} & \textbf{22.7} & \textbf{16.7} & 3.33 \\
HES-Top20 & 10.0 & 13.3 & 11.3 & 6.7 & \textbf{6.7} \\
\bottomrule
\end{tabular}
\end{table}

% ===================================================================
% RQ5: Causal Neuron Transfer
% ===================================================================
\subsection{RQ5: Causal neuron transfer}
\label{sec:rq5}

\begin{figure}[t]
  \centering
  \includegraphics[width=0.75\linewidth]{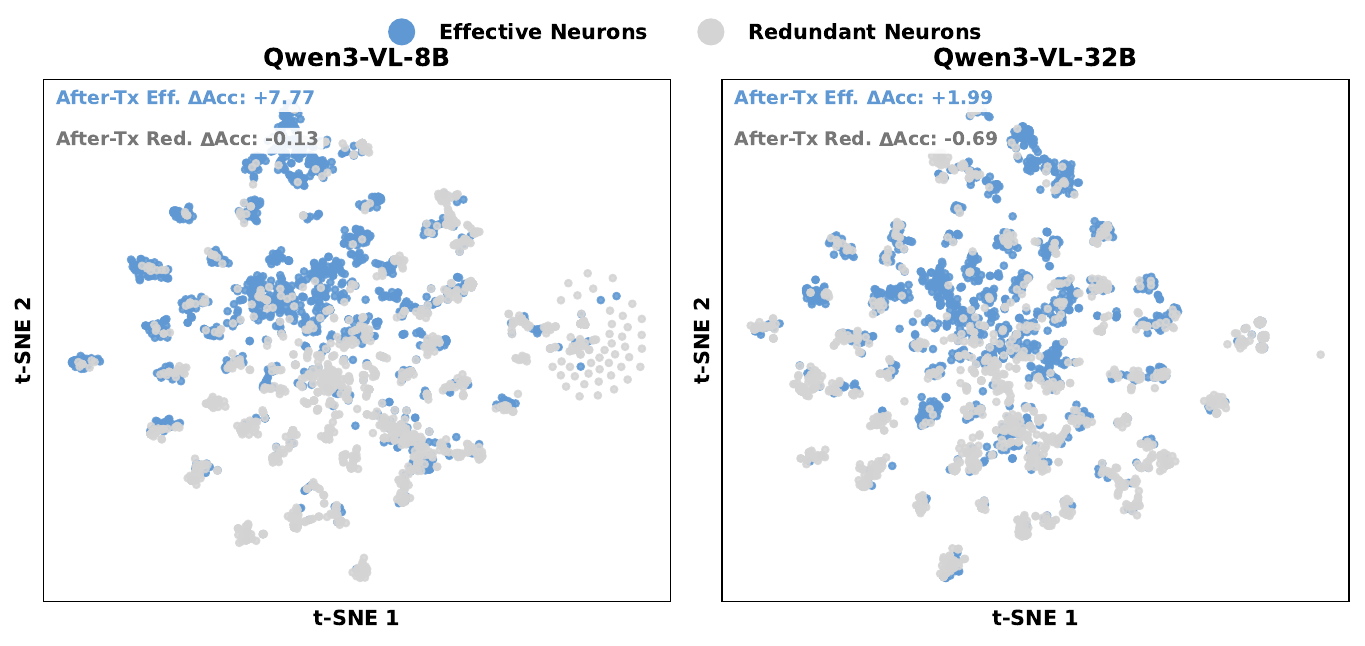}
  \caption{Causal neuron transfer on Qwen3-VL series. Each point is a neuron embedded via t-SNE on activation fingerprints. Blue: effective neurons ($w_k > 0$); gray: redundant neurons ($w_k < 0$). Transplanting effective neurons from Thinking to Instruct improves accuracy (+7.77 pp on 8B, +1.99 pp on 32B), while redundant neurons cause slight degradation.}
  \label{fig:neuron_transfer}
\end{figure}

\textbf{Setup.}
We select two model pairs (Qwen3-VL-8B and Qwen3-VL-32B), each consisting of an Instruct and Thinking variant.
For each Thinking model, we compute neuron weights $w_k$ following Section~\ref{sec:method} and partition neurons into two groups:
\emph{Effective} neurons ($w_k > 0$) associated with productive exploration, and
\emph{Redundant} neurons ($w_k < 0$) representing unproductive exploration.
\figref{fig:neuron_transfer} shows t-SNE embeddings of neuron activation, confirming that the two groups occupy distinct regions.

\textbf{Neuron transfer results.}
We transplant each neuron group from Thinking model to corresponding Instruct model by replacing FFN parameters (gate, up, and down projections) and evaluate reasoning accuracy, following activation-guided neuron transplantation~\cite{feng2026armroleconditionedneurontransplantation}.On Qwen3-VL-8B, transferring effective neurons improves accuracy by $+7.77$ pp on average, while transferring redundant ones causes $-0.13$ pp change.
On Qwen3-VL-32B, effective neurons yield $+1.99$ pp improvement versus $-0.69$ pp for redundant ones.
This asymmetry provides causal evidence that \NEX identifies neurons critical for effective reasoning.

%% file: sections/conclusion.tex
\section{Conclusion}

We introduced \NEX, a label-free neuron-dynamics scoring method that segments reasoning into E-phase vs. X-phase using a sticky two-state HMM over novelty-slope and credits neurons by their E-to-X reuse.
HMM segmentation aligns with human E-phase vs. X-phase judgments at ${\sim}90\%$ agreement (RQ1).
Raw exploration level exhibits an inverted-U relationship with accuracy; neuron weighting linearizes this into a monotonic signal by distinguishing productive from redundant exploration (RQ2).
The resulting neuron weights provide a unified signal for:
(i)~model selection under distribution shift (RQ3);
(ii)~quality CoT selection among correct answers (RQ4);
and (iii)~training-data curation with measurable student gains under equal token budgets (RQ4).
Causal transfer experiments confirm that neurons identified as effective vs. redundant by \NEX have systematic effects on reasoning performance (RQ5).
\NEX provides a practical tool for scaling reasoning pipelines when labeled validation data is scarce, requiring only a small unlabeled activation set to learn neuron weights that generalize across tasks.

\section*{Impact Statement}
\label{sec:limitations}

\NEX has several limitations.
It requires access to internal activations, which may be unavailable for closed models.
It also depends on how a reasoning trace is segmented into rows; while fixed token length works well for many CoT
formats, other formats may require more robust chunking.
In addition, our current formulation is most natural for MLP neurons with sparse activations; extending the framework to
attention heads or other internal features is an interesting direction.
Very short responses may not exhibit sufficient E-phase vs. X-phase structure for stable HMM segmentation.
The method assumes the cache mass is a faithful proxy for neuron contribution; alternative architectures may require adapting the logging point.
Finally, \NEX is a \emph{selector}, not a training method: it does not directly improve the base model and inherits the
base model's biases and knowledge limitations.
Scores are most meaningful \emph{within a model series}, for weights are trained per series.

\NEX can reduce the need for labeled evaluation when selecting models or training data, potentially lowering compute cost and enabling better deployment choices under shifting distributions.
Nevertheless, practitioners should avoid over-optimizing for narrow behavioral signatures, as these proxies may fail to capture user intent.
An extension we find particularly promising is to use \NEX as a training signal, \eg by sampling multiple CoTs and
distilling high-scoring traces, or by using the score to guide model merging and checkpoint averaging.
Another direction is to test whether the novelty-slope and HMM segmentation signals can be estimated using only black-box outputs
like token-level logprobs only, e.g. via probing or surrogate models.

%% file: sections/appendix.tex
\section{Additional Details}
\label{sec:appendix}

\subsection{Notation cheat sheet}
\label{app:notation}
\vspace{-0.25em}
\begin{center}
\footnotesize
\setlength{\tabcolsep}{4pt}
\renewcommand{\arraystretch}{1.05}
\begin{tabular}{@{}ll@{}}
\toprule
Symbol & Meaning \\
\midrule
$y=(y_1,\ldots,y_T)$ & response tokens, length $T$ \\
$r\in\{1,\ldots,R\}$ & row index ($R$ rows of $W{=}32$ tokens) \\
$T_r$ & token indices in row $r$ \\
$k$ & neuron key: $(\mathrm{layer}\ll 16)\,|\,\mathrm{unit}$ \\
$a_{k,t}\ge 0$ & activation mass of neuron $k$ at token $t$ \\
$\mathcal{N}_r$, $\mathcal{N}_{<r}$ & neurons in row $r$; cumulative before $r$ \\
$s_r$ & novelty slope (new neurons / tokens) \\
$z_r$ & processed slope (log, detrend, MAD) \\
\bottomrule
\end{tabular}
\end{center}
\vspace{-0.5em}

% We assume access to cached sparse MLP activations: for each response, we store a sparse set of activated MLP neurons along with their nonnegative activation mass.
% Each neuron is represented by a 32-bit key:
% \begin{equation}
% k = (\text{layer} \ll 16) \,|\, \text{unit}.
% \end{equation}

% \paragraph{Row granularity.}
% Given a response $y=(y_1,\dots,y_T)$, we split it into $R$ \emph{rows} (segments) $y^{(1)},\dots,y^{(R)}$ using
% fixed-size windows of $W$ tokens (we use $W=32$; Appendix~\ref{app:rows}).
% Let $T_r$ denote the set of token indices in row $r$.

% We record sparse MLP neuron activations.
% Let $a_{k,t}\ge 0$ denote the activation magnitude of neuron $k$ at token $t$, where $k$ indexes all MLP neurons across
% layers.
% To reduce storage, we keep only the top-$\kappa$ activations per token (Appendix~\ref{app:sparsity}).

\subsection{Model list}
\label{app:model-list}

\paragraph{Text models.}
\begin{itemize}[leftmargin=1.2em]
  \item \texttt{Qwen/Qwen3-4B-Instruct-2507} (Instruct endpoint)
  \item \texttt{Qwen/Qwen3-4B-Thinking-2507} (Thinking endpoint)
\end{itemize}

\paragraph{Vision-Language models (Qwen3-VL series).}
\begin{itemize}[leftmargin=1.2em]
  \item \texttt{Qwen/Qwen3-VL-4B-Instruct}, \texttt{Qwen/Qwen3-VL-4B-Thinking}
  \item \texttt{Qwen/Qwen3-VL-8B-Instruct}, \texttt{Qwen/Qwen3-VL-8B-Thinking}
  \item \texttt{Qwen/Qwen3-VL-32B-Instruct}, \texttt{Qwen/Qwen3-VL-32B-Thinking}
\end{itemize}

\paragraph{Trained model families (RQ4).}
\begin{itemize}[leftmargin=1.2em]
  \item Teacher: \texttt{Qwen/Qwen3-30B-A3B-Instruct}, \texttt{Qwen/Qwen3-30B-A3B-Thinking}
  \item Student: \texttt{Qwen/Qwen3-4B-Base} (full fine-tuning)
\end{itemize}

\subsection{Sparse activation logging}
\label{app:sparsity}

Logging full activations for all neurons and all tokens is expensive.
We uniquely represent each neuron across layers using a 32-bit key:
\begin{equation}
k = (\text{layer} \ll 16) \,|\, \text{unit}.
\end{equation}
We store a sparse representation by keeping the top-2000 MLP activations per token.
These sparse activations are then aggregated into row-level sets $\mathcal{N}_r$.
A neuron $k$ is included in $\mathcal{N}_r$ if its activation $a_{k,t} > \tau$ for any token $t$ in row $r$.
In our experiments, we use $\tau=0$ (considering all recorded top-2000 activations).
We also track response-level sums $b_k$ for further analysis.

% \subsection{Row segmentation}
% \label{app:rows}

% We use newline-delimited rows when models produce structured CoTs with explicit line breaks.
% However, to ensure robustness, when a response contains fewer than $R_{\min}$ newlines, we fallback to chunking into fixed-size windows of $W$ tokens
% (\eg $W=32$) and treat each window as a row.
% This hybrid strategy preserves semantic step boundaries when present while remaining consistent for unstructured traces.

\subsection{Novelty Slope Preprocessing}
\label{app:preprocessing}
The raw novelty slope $s_r$ (Eq.~\ref{eq:slope}) typically exhibits heavy tails and a long-term downward trend as the finite pool of neurons is exhausted. To make the signal suitable for HMM segmentation, we apply the following transformations sequentially:

\begin{enumerate}[leftmargin=1.2em]
\item \textbf{Log transform:}
\begin{equation}
\tilde{s}_r = \log(1 + s_r).
\label{eq:log-slope}
\end{equation}
\item \textbf{De-trending} to remove saturation bias:
\begin{equation}
r_t = \tilde{s}_r - (a + b\log(1+r)),
\end{equation}
where $a, b$ are fitted via least-squares regression over the trace.
\item \textbf{Robust standardization:}
\begin{equation}
z_r = \frac{r_t - \mathrm{median}(\{r_t\})}{\mathrm{MAD}(\{r_t\})+\epsilon},
\label{eq:mad-norm}
\end{equation}
where $\mathrm{MAD}$ denotes the median absolute deviation.
\end{enumerate}

% \begin{enumerate}
%     \item \textbf{Log transform:} $\tilde{s}_r = \log(1 + s_r)$ to stabilize variance.
%     \item \textbf{De-trending:} We remove the saturation bias by fitting a curve $a + b\log(1+r)$ via least-squares regression over the trace and computing residuals:
%     \begin{equation}
%     r_t = \tilde{s}_r - (a + b\log(1+r)).
%     \end{equation}
    
%     \item \textbf{Robust Standardization:} We normalize the residuals using robust statistics to mitigate outliers:
%     \begin{equation}
%     \label{eq:mad-norm}
%     z_r = \frac{r_t - \mathrm{median}(\{r_t\})}{\mathrm{MAD}(\{r_t\})+\epsilon},
%     \end{equation}
%     where $\mathrm{MAD}$ denotes the median absolute deviation.
% \end{enumerate}
The resulting $z_r$ is used as the observation sequence for the HMM.

\subsection{Details for sticky HMM segmenting}
\label{app:hmm-details}

\paragraph{Emission model fitting.}
We initialize the 2-state HMM emissions by fitting an unsupervised 2-component Gaussian mixture (diagonal covariance) to the standardized residual signal $z_r$ (Eq.~\ref{eq:mad-norm}).
Formally, the emission model is:
\begin{equation}
p(z_r \mid \text{state}=s) = \mathcal{N}(z_r; \mu_s, \sigma_s^2), \quad s \in \{0,1\},
\label{eq:hmm-emission}
\end{equation}

We use the fitted means and variances as emission parameters for Viterbi decoding and designate the state with larger emission mean as E-phase.
During decoding we use a fixed sticky transition matrix with $\rho_{hmm}=0.95$:
\begin{equation}
P(q_r = q_{r-1}) = \rho_{hmm}, \quad P(q_r \neq q_{r-1}) = 1 - \rho_{hmm}
\label{eq:hmm-transition}
\end{equation}
where $q_r$ represents hidden state.
The stickiness avoids overfitting on short traces.

\paragraph{Viterbi decoding.}
We decode the most likely state sequence using the standard Viterbi algorithm.
After decoding, we apply minimum run-length smoothing: any segment shorter than $\text{min\_run}=2$ rows is merged into its longer neighbor.
This prevents isolated single-row state flips from creating spurious E$\rightarrow$X cycles.

\paragraph{Sensitivity to initialization.}
Because we fix the mixture-model random seed and all subsequent steps are deterministic, the HMM segmentation is deterministic for a given trace.

\subsection{Complexity}
\label{app:complexity}
Computing \NEX weights requires a single forward pass per mini-set sample per candidate model, with activation logging.
Once weights $w_k$ are learned, scoring a new response requires only computing a sparse vector $b_k$ and a dot product
with $w_k$ (Eq.~\ref{eq:final-score}).
Thus, \NEX is most useful when the mini activation set is small and the candidate pool is large.

\subsection{Implementation details}
\label{app:impl}
\begin{itemize}[leftmargin=1.2em]
  \item \textbf{HMM:} Viterbi decoding via \texttt{hmmlearn.hmm.GaussianHMM}.
  \item \textbf{GMM:} Initialization via \texttt{sklearn.mixture.GaussianMixture} (fixed seed).
  \item \textbf{Cache row size:} ${\sim}32$ tokens per row.
  \item \textbf{HMM segmentation:} 2-state Gaussian emission, sticky transition with $\rho=0.95$, minimum run length $=2$.
  \item \textbf{Novelty slope preprocessing:} new-neuron count per row; $\log(1+s_r)$; de-trending by $a+b\log(1+r)$; MAD normalization (Appendix~\ref{app:preprocessing}).
  \item \textbf{Strength gating:} include an E$\rightarrow$X cycle iff $\mathrm{median}(s_E) > \mathrm{median}(s)$ within the run (Eq.~\ref{eq:reuse-strength}).
  \item \textbf{Weight formula:} $w_k=\tanh\!\big(\log\big((m_k^{+}+\epsilon)/(m_k^{-}+\epsilon)\big)\big)$ with $\epsilon=10^{-6}$ (Eq.~\ref{eq:weight}).
  \item \textbf{Score (Good-Mass Fraction):} per run $\sum_k b_k[w_k]_+ / \sum_k b_k|w_k|$ (0 if denom$=0$), then mean over runs (Eq.~\ref{eq:final-score}).
  \item \textbf{Self-calibration:} by default, each candidate trains its own weights from its own mini cache; an ablation uses shared baseline weights.
\end{itemize}

% \subsection{NEX algorithm pseudocode}
% \label{app:code}
% \begin{algorithm}[H]
% \caption{\NEX}
% \label{alg:reusehmm}
% \begin{algorithmic}[1]
% \REQUIRE Mini activation set $\{x^{(j)}\}_{j=1}^{N}$, candidate models $\{\theta\}$
% \STATE Run the Thinking model on the mini-set; record sparse activations $a_{k,t}$ and compute novelty slopes $s_r$ (Eq.~\eqref{eq:slope})
% \STATE Log-transform, de-trend, and MAD-normalize slopes (Appendix~\ref{app:preprocessing}); segment each run into E-phase/X-phase via sticky 2-state HMM (Eq.~\eqref{eq:hmm-transition})
% \STATE For each E$\rightarrow$X pair, compute reuse-based progress (Eq.~\eqref{eq:reuse-progress}) and consolidation (Eq.~\eqref{eq:reuse-cons})
% \STATE Accumulate neuron masses $m_k^{+},m_k^{-}$ on newly introduced neurons and compute weights $w_k$ (Eq.~\eqref{eq:weight})
% \FOR{each new response $y$}
%   \STATE Compute sparse response-level masses $b_k(y)$ and $\mathrm{Score}(y)=\sum_k b_k[w_k]_+ / \sum_k b_k|w_k|$
% \ENDFOR
% \end{algorithmic}
% \end{algorithm}

\subsection{Human audit details and error analysis}
\label{app:human}

\paragraph{Annotation guidelines.}
Two independent annotators label each segmented block using the following criteria:
\begin{itemize}[leftmargin=1.2em]
  \item \textbf{E-phase (exploration):} presence of reflective markers (``wait'', ``but'', ``maybe''), error correction or backtracking, re-interpretation of the problem statement, proposing alternative approaches, case analysis, or forming hypotheses to guide subsequent verification.
  \item \textbf{X-phase (exploitation):} sustained progress under a prior conclusion or framework, chains of equalities or modular arithmetic, extensive numerical computation, explicit substitution/simplification/solving steps, logical extension of a core hypothesis into corollaries, or consolidating all conclusions toward a final answer.
\end{itemize}

\paragraph{Example cases.}
Below we show representative segments where human labels agree with HMM segmentation.

\textbf{E-phase (exploration):}
\begin{quote}
\small
\textit{``Wait, but let's check with the n=3 case to be sure. For n=3, truck bottom to top [B1,B2,B3], second from bottom is B2. Probability B2 is bottom of shed (Y1=B2) is 1/4, as we calculated. For n=3, (1/2)\^{}(3-1)=1/4, which matches. For n=2, truck [B1,B2], second from bottom is B2, probability Y1=B2 is 1/2=(1/2)\^{}(2-1), correct. So for general n, the probability that the second from the bottom in the truck is the bottom in the shed is (1/2)\^{}(n-1). Wait,...''}
\end{quote}
This segment exhibits typical E-phase markers: ``Wait, but let's check'' indicates verification/backtracking, and the model tests multiple cases (n=3, n=2) before generalizing.

\textbf{X-phase (exploitation):}
\begin{quote}
\small
\textit{``...the large cube is partially painted with gray paint, as shown in the provided Asymptote (asy) diagram. From the asy code, there are 12 gray regions drawn across the diagram. Each region corresponds to a painted unit face on one of the six faces of the large cube. This implies that each face of the large cube has 12 painted unit faces, and thus 13 unpainted unit faces per face (since each face has 25 unit faces total).''}
\end{quote}
This segment exhibits typical X-phase markers: sustained derivation without hesitation, logical extension from observation to conclusion, and numerical computation.

\paragraph{Novelty slope vs entropy disagreement cases.}
\label{app:novelty-entropy}
We identify three patterns of disagreement between entropy and novelty slope.
Below are representative examples where novelty slope more accurately captures exploration behavior.

\textbf{Type A: Zero entropy but exploring (enumeration degradation).}
\begin{quote}
\small
\texttt{0  \textbackslash n+16 → 26  \textbackslash n+22 → 48  \textbackslash n+64 → 112  \textbackslash n+93 →}
\end{quote}
This segment has near-zero entropy ($H \approx 10^{-7}$) because the model generates deterministic arithmetic sequences.
However, novelty slope remains elevated ($s_r = 0.09$) because the model explores different accumulation paths---each step recruits new computational circuits.
Entropy classifies this as exploitation; novelty slope correctly identifies it as exploration.

\textbf{Type B: Low entropy but semantically exploring (confident strategy switch).}
\begin{quote}
\small
\texttt{to \$ 2\^{}b - 1 \$) that have more 1s than 0s.\textbackslash n\textbackslash nNote: \$ b \$-bit numbers must have}
\end{quote}
This segment has low mean entropy ($H = 0.23$) because the model confidently introduces a new definition.
Yet novelty slope is elevated ($s_r = 1.4$) because the model establishes a new reasoning framework.
The phrase ``Note: $b$-bit numbers must have'' signals a conceptual pivot that entropy misses.

\textbf{Type C: High entropy but not exploring (stylistic choice).}
\begin{quote}
\small
\texttt{is orthogonal, and thus \$ \textbackslash mathbf\{P\}\^{}T = \textbackslash mathbf\{P\}\^{}\{-1\} \$.\textbackslash n\textbackslash nSo:\textbackslash n\$\$}
\end{quote}
This segment has elevated entropy ($H_{\max} = 1.86$) because connectors like ``and thus'' and ``So'' have multiple paraphrases.
However, novelty slope is low ($s_r = 0.53$) because the underlying derivation reuses established circuits.
Entropy would flag this as exploration; novelty slope correctly identifies it as exploitation.

\subsection{Phase-level entropy comparison}
\label{app:entropy-comparison}

We compute mean token entropy within E-phase and X-phase segments across all merged model families.
\tabref{tab:entropy_phase} confirms that E-phase consistently exhibits higher entropy than X-phase, supporting the interpretation that exploration involves more diverse token distributions.

\begin{table}[h]
\centering
\caption{Mean token entropy by phase for merged models. E-phase shows consistently higher entropy across all model families.}
\label{tab:entropy_phase}
\small
\setlength{\tabcolsep}{6pt}
\begin{tabular}{@{}lcc@{}}
\toprule
Model Family & E-phase Entropy & X-phase Entropy \\
\midrule
Qwen3-4B & 0.2475 & 0.1688 \\
Qwen3-VL-4B & 0.2825 & 0.1852 \\
Qwen3-VL-8B & 0.2917 & 0.1813 \\
Qwen3-VL-32B & 0.2829 & 0.1711 \\
\midrule
Average & 0.2765 & 0.1767 \\
\bottomrule
\end{tabular}
\end{table}

\subsection{Effective vs.\ redundant exploration cases}
\label{app:effective-redundant}

We present representative cases illustrating the distinction between effective and redundant exploration within the same reasoning trace.
The key metrics are:
\emph{reuse share} (fraction of X-phase activation mass from neurons introduced in E-phase) and
\emph{consolidation} (degree of novelty slope reduction from E to X).
Effective exploration satisfies both progress $>0$ and consolidation $>0$; redundant exploration fails one or both criteria.

\paragraph{Problem: 5$\times$5$\times$5 cube painting.}
\emph{``A solid $5\times 5\times 5$ cube is composed of unit cubes. Each face of the large cube is partially painted with gray paint...''}

\textbf{Effective exploration} (reuse share $= 0.74$, consolidation $= 0.83$):
\begin{quote}
\small
\texttt{y \textbackslash{} x | 1 | 2 | 3 | 4 | 5}\\
\texttt{1~~~| P | N | N | N | P}\\
\texttt{2~~~| N | P | P | P | N}\\
\texttt{3~~~| N | P | N | P | N}\\
\texttt{4~~~| N | P | P | P | N}\\
\texttt{5~~~| P | N | N | N | P}\\[0.5em]
\texttt{Yes! This is a 5x5 grid where P means the front face of cube...}
\end{quote}
The model constructs a structured P/N table to analyze painting patterns.
This framework is \emph{reused extensively} in subsequent derivations (74\% of X-phase activation mass comes from neurons introduced here).

\textbf{Redundant exploration} (reuse share $= 0$, consolidation $= 0.26$):
\begin{quote}
\small
\texttt{Perfect, that's 1 + 3 + 2 + 3 + 2 = 11? Wait, no: row 0:1, row1:3, row2:2, row3:3, row4:2 → total 11, but we have 12 fills. Wait...}
\end{quote}
The model attempts a counting approach but immediately discovers an error (``11? Wait, no'').
This attempt is \emph{discarded} without contributing to subsequent reasoning (reuse share $= 0$).

\paragraph{Problem: Binary representation counting.}
\emph{``Find the number of positive integers $\le 2003$ whose base-2 representation has more 1s than 0s...''}

\textbf{Effective exploration} (reuse share $= 0.09$, consolidation $= 0.86$):
\begin{quote}
\small
\texttt{L=3: Numbers: 4-7: "100"(t=1), "101"(2), "110"(2), "111"(3)}\\
\texttt{t > 1.5} $\Rightarrow$ \texttt{t>=2: t=2,3: 3 numbers}\\
\texttt{Count=3}
\end{quote}
The model establishes a systematic enumeration framework by bit-length, which is reused for subsequent cases.

\textbf{Redundant exploration} (reuse share $= 0$, consolidation $= 0.32$):
\begin{quote}
\small
\texttt{from 5 to 9=256, which is half of 512, makes sense for odd n=9, symmetric sums)}\\
\texttt{Wait, let's check L=7 again...}
\end{quote}
The model backtracks to re-verify an earlier case (``Wait, let's check L=7 again'') without introducing reusable structure.

\paragraph{Implication for neuron weighting.}
These cases demonstrate why raw exploration counts are insufficient:
both effective and redundant exploration occur within the same run.
\NEX assigns $w_k > 0$ to neurons in effective cycles and $w_k < 0$ to those in redundant cycles,
enabling the score to distinguish productive reasoning from wasted computation.

\subsection{Extended inverted-U analysis across model families}
\label{app:inverted-u-extended}

\begin{figure}[h]
  \centering
  \includegraphics[width=0.8\linewidth]{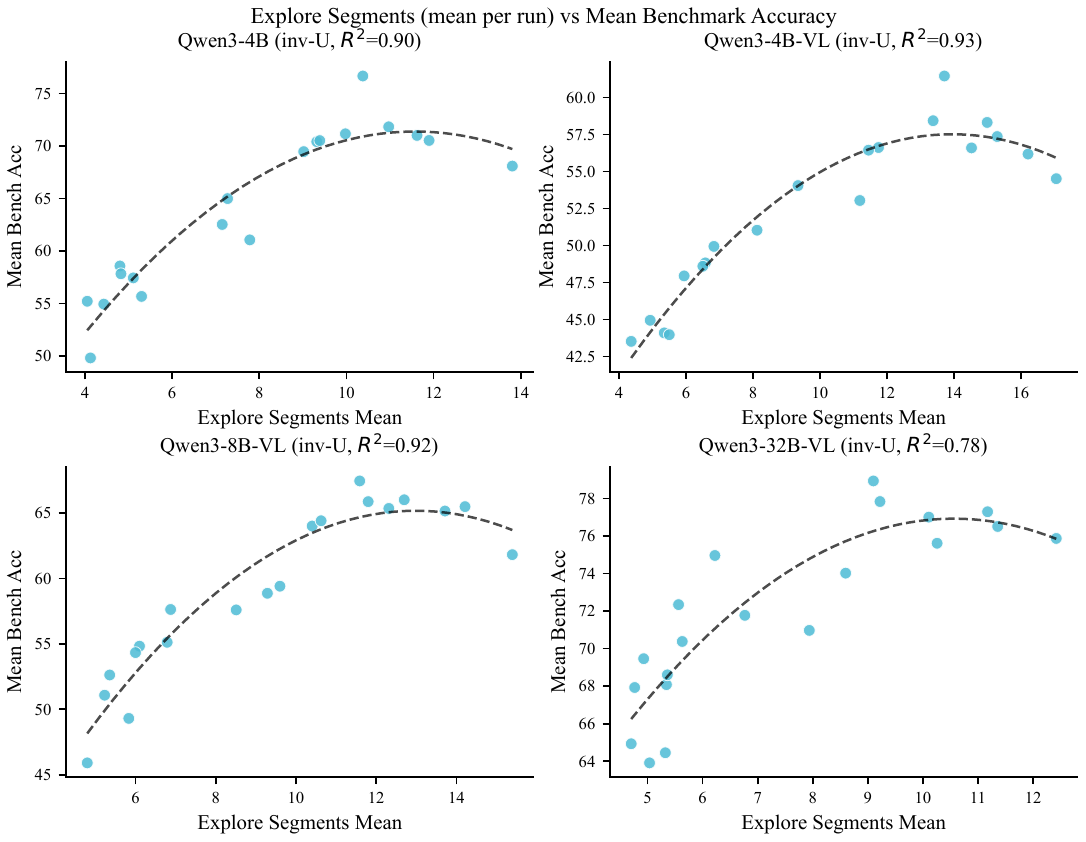}
  \caption{Inverted-U relationship between E-phase segment count and accuracy across four model series. Each panel shows one model family; each point is a merged variant. The non-monotonic pattern indicates an optimal exploration level exists for each series.}
  \label{fig:inverted_u_extended}
\end{figure}

\figref{fig:inverted_u_extended} shows the inverted-U relationship between raw exploration level and accuracy across four model series (Qwen3-4B, Qwen3-VL-4B, Qwen3-VL-8B, Qwen3-VL-32B).
For each model, we plot the mean number of HMM-detected E-phase segments on the mini trajectories against the mean accuracy on five downstream benchmarks (AIME24, AIME25, GPQA, HMMT25, BRUMO25).
Each point corresponds to one merged model variant.
The relationship is generally non-monotonic: accuracy increases as models exhibit more exploratory phase switches, but saturates (and can decline) when exploration becomes overly fragmented, consistent with a ``too little vs.\ too much exploration'' trade-off.

\subsection{Best-of-$n$ CoT quality analysis}
\label{app:best-of-n-cot}

We analyze the qualitative differences between \NEX-selected CoTs (Best-of-3 by \NEX score) and randomly sampled CoTs (run 1) on problems where both produce correct answers.
This isolates reasoning quality differences from correctness.

\paragraph{Quantitative patterns.}
Across multiple model variants, \NEX-selected CoTs consistently exhibit:
\begin{itemize}[leftmargin=1.2em]
  \item \textbf{Fewer backtracking markers:} reduced occurrences of ``Wait'', ``But wait'', and similar self-correction phrases.
  \item \textbf{Shorter token length:} more concise reasoning under equivalent correctness.
  \item \textbf{Clearer structure:} faster progression to key derivations with less preamble.
\end{itemize}

\paragraph{Case A: Reduced backtracking.}
The random CoT begins with extensive preamble (``This is a complex or challenging question...'') and contains 135 ``wait'' markers with 9 ``but wait'' instances, reflecting repeated self-correction.
The \NEX-selected CoT proceeds more directly to the state-machine formulation, reducing ``wait'' markers to 61 and eliminating ``but wait'' entirely, while achieving the same correct answer in fewer tokens ($21716 \to 17345$).

\paragraph{Case B: Higher information density.}
For a Bernoulli sufficient statistic problem, the random CoT provides textbook-style exposition of sufficient statistics definitions before solving (${\sim}16916$ tokens).
The \NEX-selected CoT directly writes the joint pmf, identifies the sufficient statistic $S = X_1 + X_2 + X_3$, and checks options (${\sim}4193$ tokens)---a 4$\times$ reduction while maintaining correctness.

\paragraph{Implication.}
These patterns support the interpretation that \NEX preferentially selects CoTs exhibiting exploit-phase convergence: fewer exploratory detours, more direct derivation paths, and higher reasoning efficiency per token.

\subsection{Rollout data pipeline (RQ4)}
\label{app:rollout-data}

\paragraph{Source datasets.}
We use three public math reasoning datasets:
\begin{itemize}[leftmargin=1.2em]
  \item \texttt{dapo-math-17k}: 17,917 problems
  \item \texttt{deepmath-103k}: 101,882 problems (with difficulty labels 1.0--10.0)
  \item \texttt{deepscaler}: 40,315 problems
\end{itemize}
Total: 160,114 problems.

\paragraph{Rollout generation.}
We generate CoT rollouts from two endpoint models:
\begin{itemize}[leftmargin=1.2em]
  \item \texttt{Qwen3-30B-A3B-Instruct}: instruction-tuned variant (under-explores)
  \item \texttt{Qwen3-30B-A3B-Thinking}: thinking-tuned variant (over-explores)
\end{itemize}
Generation uses vLLM with 8$\times$H20 GPUs, temperature $T{=}0.7$, top-$p{=}0.9$, and max tokens $32{,}768$.

\paragraph{Label generation.}
For each problem, we record a 2-bit binary label:
\begin{itemize}[leftmargin=1.2em]
  \item Bit 0: Instruction model correctness ($1{=}$correct)
  \item Bit 1: Thinking model correctness ($1{=}$correct)
\end{itemize}
We also record a finish label ($1{=}$normal stop, $0{=}$truncated).

\subsection{Implementation checklist for reproduction}
\label{app:checklist}
To reproduce the method:
\begin{enumerate}[leftmargin=1.2em]
  \item Choose two endpoints (Instruct vs Thinking) and generate merged or trained-mixture model variants.
  \item Construct a mini activation set (100 problems) and run all candidate model variants on it.
  \item Record sparse MLP activations (top-2000 per token) and compute novelty slopes $s_r$.
  \item Log-transform, de-trend, and MAD-normalize slopes; fit sticky 2-state HMM; decode E-phase/X-phase via Viterbi.
  \item Compute reuse-based progress, consolidation, and strength; accumulate $m_k^+,m_k^-$ and compute weights $w_k$.
  \item Score new model variants by their mean \NEX score on the mini set; score new samples by per-sample \NEX score.
\end{enumerate}

%% file: references.bib
@misc{
anonymous2026unified,
title={Unified Data Selection for {LLM} Reasoning},
author={Anonymous},
year={2026},
url={https://openreview.net/forum?id=heVn5cNfje}
}

@misc{aime24,
  title        = {AIME 2024 Benchmark},
  author       = {Zhang, Di and Math-AI},
  year         = {2024},
  url          = {https://huggingface.co/datasets/HuggingFaceH4/aime_2024},
  note         = {American Invitational Mathematics Examination 2024}
}

@misc{aime25,
  title        = {AIME 2025 Benchmark},
  author       = {Zhang, Di and Math-AI},
  year         = {2025},
  url          = {https://huggingface.co/datasets/math-ai/aime25},
  note         = {American Invitational Mathematics Examination 2025}
}

@misc{hmmt25,
  title        = {HMMT 2025 Benchmark},
  author       = {MathArena},
  year         = {2025},
  url          = {https://huggingface.co/datasets/MathArena/hmmt_feb_2025},
  note         = {Harvard--MIT Mathematics Tournament 2025}
}

@misc{brumo25,
  title        = {BRUMO 2025 Benchmark},
  author       = {MathArena},
  year         = {2025},
  url          = {https://huggingface.co/datasets/MathArena/brumo_2025},
  note         = {British University Mathematical Olympiad 2025}
}

@misc{feng2026armroleconditionedneurontransplantation,
      title={ARM: Role-Conditioned Neuron Transplantation for Training-Free Generalist LLM Agent Merging}, 
      author={Zhuoka Feng and Kang Chen and Sihan Zhao and Kai Xiong and Yaoning Wang and Minshen Yu and Junjie Nian and Changyi Xiao and Yixin Cao and Yugang Jiang},
      year={2026},
      eprint={2601.07309},
      archivePrefix={arXiv},
      primaryClass={cs.AI},
      url={https://arxiv.org/abs/2601.07309}, 
}

@article{wei2022chain,
  title        = {Chain-of-Thought Prompting Elicits Reasoning in Large Language Models},
  author       = {Wei, Jason and Wang, Xuezhi and Schuurmans, Dale and Bosma, Maarten and Ichter, Brian and Xia, Fei and Chi, Ed and Le, Quoc and Zhou, Denny},
  journal      = {Advances in Neural Information Processing Systems},
  year         = {2022},
  url          = {https://arxiv.org/abs/2201.11903}
}

@inproceedings{wang2023selfconsistency,
  title        = {Self-Consistency Improves Chain of Thought Reasoning in Language Models},
  author       = {Wang, Xuezhi and Wei, Jason and Schuurmans, Dale and Le, Quoc and Chi, Ed and Narang, Sharan and Chowdhery, Aakanksha and Zhou, Denny},
  booktitle    = {International Conference on Learning Representations (ICLR)},
  year         = {2023},
  url          = {https://arxiv.org/abs/2203.11171}
}

@article{shannon1948communication,
  title        = {A Mathematical Theory of Communication},
  author       = {Shannon, Claude E.},
  journal      = {The Bell System Technical Journal},
  volume       = {27},
  number       = {3},
  pages        = {379--423},
  year         = {1948}
}

@book{sutton2018rl,
  title        = {Reinforcement Learning: An Introduction},
  author       = {Sutton, Richard S. and Barto, Andrew G.},
  edition      = {2},
  publisher    = {MIT Press},
  year         = {2018},
  url          = {http://incompleteideas.net/book/the-book-2nd.html}
}

@article{wortsman2022soups,
  title        = {Model Soups: Averaging Weights of Multiple Fine-Tuned Models Improves Accuracy Without Increasing Inference Time},
  author       = {Wortsman, Mitchell and Ilharco, Gabriel and Gadre, Samir Y. and Roelofs, Rebecca and Gontijo-Lopes, Raphael and Morcos, Ari S. and Namkoong, Hongseok and Farhadi, Ali and Carmon, Yair and Kornblith, Simon and Schmidt, Ludwig},
  journal      = {arXiv preprint arXiv:2203.05482},
  year         = {2022},
  url          = {https://arxiv.org/abs/2203.05482}
}

@article{yadav2023ties,
  title        = {{TIES-Merging}: Resolving Interference When Merging Models},
  author       = {Yadav, Prateek and Tam, Derrick and Choshen, Leshem and Raffel, Colin and Bansal, Mohit},
  journal      = {arXiv preprint arXiv:2306.01708},
  year         = {2023},
  url          = {https://arxiv.org/abs/2306.01708}
}

@article{zhang2025ttssurvey,
  title        = {A Survey on Test-Time Scaling in Large Language Models: What, How, Where, and How Well?},
  author       = {Zhang, Qiyuan and Lyu, Fuyuan and Sun, Zexu and Wang, Lei and Zhang, Weixu and Hua, Wenyue and Wu, Haolun and Guo, Zhihan and Wang, Yufei and Muennighoff, Niklas and King, Irwin and Liu, Xue and Ma, Chen},
  journal      = {arXiv preprint arXiv:2503.24235},
  year         = {2025},
  url          = {https://arxiv.org/abs/2503.24235}
}

@inproceedings{snell2025ttcoptimal,
  title        = {Scaling {LLM} Test-Time Compute Optimally Can be More Effective than Scaling Parameters for Reasoning},
  author       = {Snell, Charlie Victor and Lee, Jaehoon and Xu, Kelvin and Kumar, Aviral},
  booktitle    = {International Conference on Learning Representations (ICLR)},
  year         = {2025},
  url          = {https://openreview.net/forum?id=4FWAwZtd2n}
}

@article{wang2025beyond8020,
  title   = {Beyond the 80/20 Rule: High-Entropy Minority Tokens Drive Effective Reinforcement Learning for {LLM} Reasoning},
  author  = {Wang, Shenzhi and Yu, Le and Gao, Chang and Zheng, Chujie and Liu, Shixuan and Lu, Rui and Dang, Kai and Chen, Xionghui and Yang, Jianxin and Zhang, Zhenru and Liu, Yuqiong and Yang, An and Zhao, Andrew and Yue, Yang and Song, Shiji and Yu, Bowen and Huang, Gao and Lin, Junyang},
  journal = {arXiv preprint arXiv:2506.01939},
  year    = {2025},
  url     = {https://arxiv.org/abs/2506.01939}
}

@article{hitit2025mergingstudy,
  title        = {A Systematic Study of Model Merging Techniques in Large Language Models},
  author       = {Hitit, O\u{g}uz Ka\u{g}an and Girrbach, Leander and Akata, Zeynep},
  journal      = {arXiv preprint arXiv:2511.21437},
  year         = {2025},
  url          = {https://arxiv.org/abs/2511.21437}
}

@article{rahamim2026mergeability,
  title        = {Will it Merge? On The Causes of Model Mergeability},
  author       = {Rahamim, Adir and Yehudai, Asaf and Carmeli, Boaz and Choshen, Leshem and Mass, Yosi and Belinkov, Yonatan},
  journal      = {arXiv preprint arXiv:2601.06672},
  year         = {2026},
  url          = {https://arxiv.org/abs/2601.06672}
}

@article{kadavath2022lmsknow,
  title   = {Language Models (Mostly) Know What They Know},
  author  = {Kadavath, Saurav and Conerly, Tom and Askell, Amanda and Henighan, Tom and Drain, Dawn and Perez, Ethan and Bowman, Samuel R. and Kaplan, Jared and others},
  journal = {arXiv preprint arXiv:2207.05221},
  year    = {2022},
  url     = {https://arxiv.org/abs/2207.05221}
}

@article{rein2023gpqa,
  title   = {GPQA: A Graduate-Level Google-Proof Q\&A Benchmark},
  author  = {Rein, David and Hou, Betty Li and Stickland, Asa Cooper and Petty, Jackson and Pang, Richard Yuanzhe and Dirani, Julien and Michael, Julian and Bowman, Samuel R.},
  journal = {arXiv preprint arXiv:2311.12022},
  year    = {2023},
  url     = {https://arxiv.org/abs/2311.12022}
}

@article{skean2025layerbylayer,
  title   = {Layer by Layer: Uncovering Hidden Representations in Language Models},
  author  = {Skean, Oscar and Arefin, Md Rifat and Zhao, Dan and Patel, Niket and Naghiyev, Jalal and LeCun, Yann and Shwartz-Ziv, Ravid},
  journal = {arXiv preprint arXiv:2502.02013},
  year    = {2025},
  url     = {https://arxiv.org/abs/2502.02013}
}

@article{chen2025llmssignaltheyreright,
  title   = {Do {LLM}s Signal When They're Right? Evidence from Neuron Agreement},
  author  = {Chen, Kang and Wang, Yaoning and Xiong, Kai and Feng, Zhuoka and Sun, Wenhe and Chen, Haotian and Cao, Yixin},
  journal = {arXiv preprint arXiv:2510.26277},
  year    = {2025},
  url     = {https://arxiv.org/abs/2510.26277}
}

@article{lewkowycz2022minerva,
  title   = {Solving Quantitative Reasoning Problems with Language Models},
  author  = {Lewkowycz, Aitor and Andreassen, Anders and Dohan, David and Dyer, Ethan and Michalewski, Henryk and Ramasesh, Vinay and Slone, Ambrose and Anil, Cem and Schlag, Imanol and Gutman-Solo, Theo and Wu, Yuhuai and Neyshabur, Behnam and Gur-Ari, Guy and Misra, Vedant},
  journal = {arXiv preprint arXiv:2206.14858},
  year    = {2022},
  url     = {https://arxiv.org/abs/2206.14858}
}

@article{welleck2024metageneration,
  title   = {From Decoding to Meta-Generation: Inference-Time Algorithms for Large Language Models},
  author  = {Welleck, Sean and Bertsch, Amanda and Finlayson, Matthew and Schoelkopf, Hailey and Xie, Alex and Neubig, Graham and Kulikov, Ilia and Harchaoui, Zaid},
  journal = {arXiv preprint arXiv:2406.16838},
  year    = {2024},
  url     = {https://arxiv.org/abs/2406.16838}
}

@article{wu2024inferencescaling,
  title   = {Inference Scaling Laws: An Empirical Analysis of Compute-Optimal Inference for Problem-Solving with Language Models},
  author  = {Wu, Yangzhen and Sun, Zhiqing and Li, Shanda and Welleck, Sean and Yang, Yiming},
  journal = {arXiv preprint arXiv:2408.00724},
  year    = {2024},
  url     = {https://arxiv.org/abs/2408.00724}
}

@article{yao2023tree,
  title   = {Tree of Thoughts: Deliberate Problem Solving with Large Language Models},
  author  = {Yao, Shunyu and Yu, Dian and Zhao, Jeffrey and Shafran, Izhak and Griffiths, Thomas L. and Cao, Yuan and Narasimhan, Karthik},
  journal = {arXiv preprint arXiv:2305.10601},
  year    = {2023},
  url     = {https://arxiv.org/abs/2305.10601}
}

@article{kuhn2023semantic,
  title   = {Semantic Uncertainty: Linguistic Invariances for Uncertainty Estimation in Natural Language Generation},
  author  = {Kuhn, Lorenz and Gal, Yarin and Farquhar, Sebastian},
  journal = {arXiv preprint arXiv:2302.09664},
  year    = {2023},
  url     = {https://arxiv.org/abs/2302.09664}
}

@article{holtzman2020curious,
  title   = {The Curious Case of Neural Text Degeneration},
  author  = {Holtzman, Ari and Buys, Jan and Du, Li and Forbes, Maxwell and Choi, Yejin},
  journal = {arXiv preprint arXiv:1904.09751},
  year    = {2019},
  url     = {https://arxiv.org/abs/1904.09751}
}

@article{cao2025modelutilitylawevaluating,
  title   = {Model Utility Law: Evaluating {LLM}s beyond Performance through Mechanism Interpretable Metric},
  author  = {Cao, Yixin and Ying, Jiahao and Wang, Yaoning and Qiu, Xipeng and Huang, Xuanjing and Jiang, Yugang},
  journal = {arXiv preprint arXiv:2504.07440},
  year    = {2025},
  url     = {https://arxiv.org/abs/2504.07440}
}

@article{ilharco2023taskarithmetic,
  title   = {Editing Models with Task Arithmetic},
  author  = {Ilharco, Gabriel and Ribeiro, Marco Tulio and Wortsman, Mitchell and Gururangan, Suchin and Schmidt, Ludwig and Hajishirzi, Hannaneh and Farhadi, Ali},
  journal = {arXiv preprint arXiv:2212.04089},
  year    = {2023},
  url     = {https://arxiv.org/abs/2212.04089}
}

@article{matena2022fishermerging,
  title   = {Merging Models with Fisher-Weighted Averaging},
  author  = {Matena, Michael and Raffel, Colin},
  journal = {arXiv preprint arXiv:2111.09832},
  year    = {2022},
  url     = {https://arxiv.org/abs/2111.09832}
}
